\documentclass[10pt, conference]{IEEEtran}
\IEEEoverridecommandlockouts
\usepackage{cite}
\usepackage{amsmath,amssymb,amsfonts}
\usepackage{algorithmic}
\usepackage{graphicx}
\usepackage{textcomp}
\usepackage{xcolor}
\usepackage{times}
\usepackage{bm}
\usepackage{comment}
\usepackage{subcaption}
\usepackage{booktabs}
\usepackage{makecell}
\usepackage[export]{adjustbox}
\newcommand{\red}[1]{{\color{red}#1}}

\def\BibTeX{{\rm B\kern-.05em{\sc i\kern-.025em b}\kern-.08em
    T\kern-.1667em\lower.7ex\hbox{E}\kern-.125emX}}
\begin{document}
\title{\LARGE \bf Learning-Based Collaborative Control for\\ Bi-Manual Tactile-Reactive Grasping 

}
\author{Leonel Giacobbe$^{1}$, Jingdao Chen$^{1}$, Chuangchuang Sun$^{2}$
\thanks{$^{1}$ Department of Computer Science, Mississippi State University. Starkville, MS, USA}%
\thanks{$^{2}$ Department of Mechanical Engineering, Villanova University. Villanova, PA, USA}%
}

\maketitle

\begin{abstract}
Grasping is a core task in robotics with various applications. However, most current implementations are primarily designed for rigid items, and their performance drops considerably when handling fragile or deformable materials that require real-time feedback. Meanwhile, tactile-reactive grasping focuses on a single agent, which limits their ability to grasp and manipulate large, heavy objects. 
To overcome this, we propose a learning-based, tactile-reactive multi-agent Model Predictive Controller (MPC) for grasping a wide range of objects with different softness and shapes, beyond the capabilities of preexisting single-agent implementations. Our system uses two Gelsight Mini tactile sensors \cite{yuan2017gelsight} to extract real-time information on object texture and stiffness. This rich tactile feedback is used to estimate contact dynamics and object compliance in real time, enabling the system to adapt its control policy to diverse object geometries and stiffness profiles. The learned controller operates in a closed loop, leveraging tactile encoding to predict grasp stability and adjust force and position accordingly.
Our key technical contributions include a multi-agent MPC formulation trained on real contact interactions, a tactile-data driven method for inferring grasping states, and a coordination strategy that enables collaborative control.
By combining tactile sensing and a learning-based multi-agent MPC, our method offers a robust, intelligent solution for collaborative grasping in complex environments, significantly advancing the capabilities of multi-agent systems.
Our approach is validated through extensive experiments against independent PD and MPC baselines. Our pipeline outperforms the baselines regarding success rates in achieving and maintaining stable grasps across objects of varying sizes and stiffness.

\end{abstract}

\section{Introduction}
Robotic grasping and manipulation have been the subject of extensive research, with many approaches promising to be valid methods to enhance accuracy in both initial grasp success and the ability to maintain stable grips\cite{zhang2022roboticgraspingclassicalmodern}. However, the majority of these methods have focused on rigid objects, often overlooking the challenges posed by soft or irregularly shaped items. Grasping and manipulating such objects remains difficult due to their compliance, variable surface properties, and the need to apply finely tuned contact forces.

Humans naturally rely on tactile feedback to infer properties such as stiffness, texture, and rigidity, which are crucial information when handling objects. In robotics, tactile sensors offer the potential to replicate this capability, providing dense, high-dimensional data well-suited for learning-based controllers. The rich sensory information embedded in tactile images can be effectively leveraged by neural networks to improve performance in grasping and manipulation in commonplace robotic scenarios.

\begin{figure}
\centering
    \includegraphics[width=1.0\linewidth, clip]{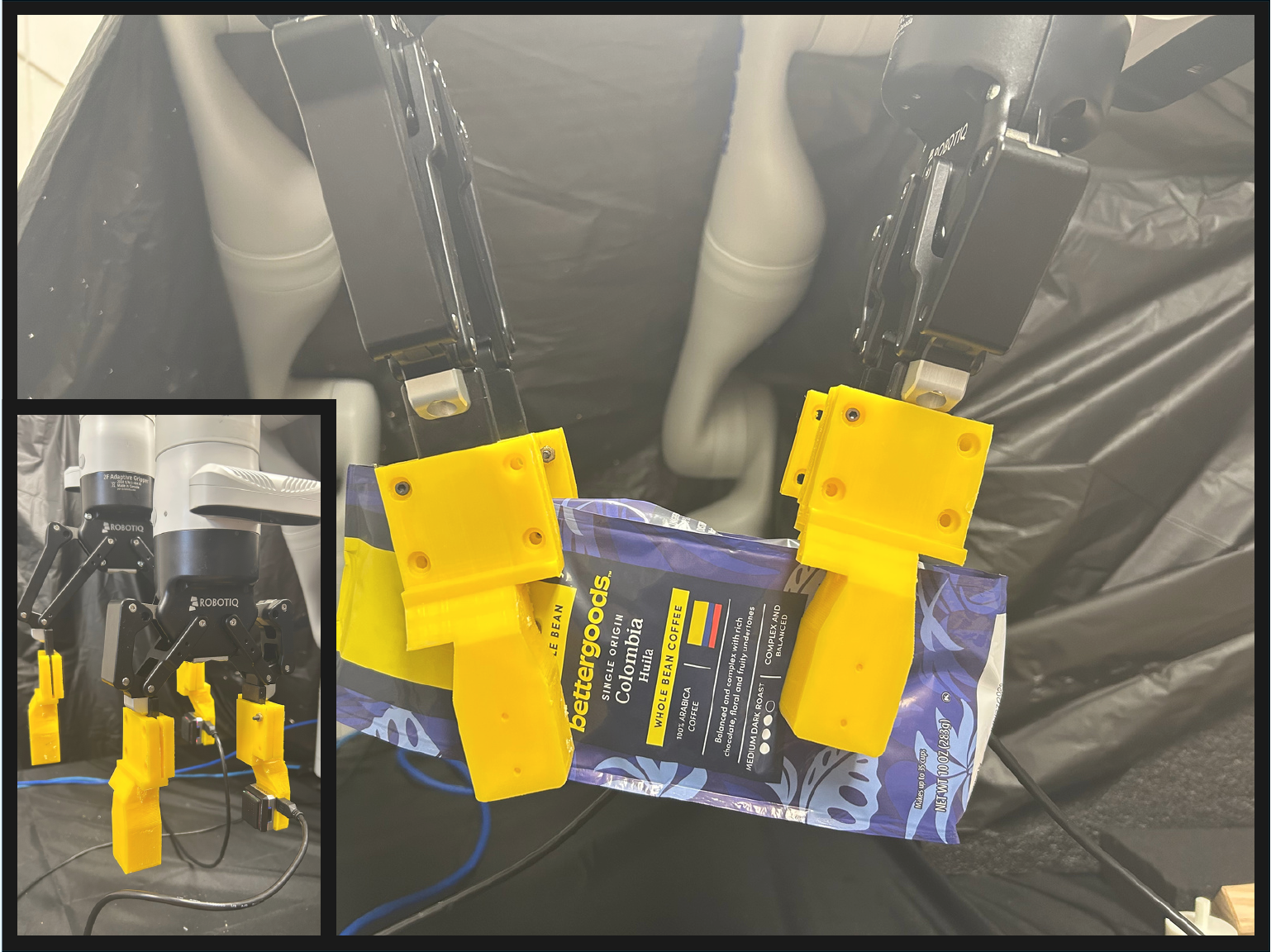}%
\caption{A bag of coffee beans grasped by two robotic grippers mounted with Gelsight tactile sensors. 
}
\label{fig:arms_grasping}
\end{figure}

Despite progress in single-agent tactile control, effective multi-agent grasping coordination remains a fundamental challenge in manipulation tasks, especially in unstructured, cluttered environments.
Recent work, such as LeTac-MPC \cite{xuletac2024}, proposed a learning-based Model Predictive Control (MPC) layer for tactile-reactive grasping using a single robotic manipulator. While promising, their implementation assumes ideal object positioning and orientation, which may not be a realistic expectation in real-world environments. In real-world deployments, where grasp prediction algorithms such as \cite{sundermeyer2021contact} are employed, performance can decrease significantly, as some objects can be quite challenging to grasp and manipulate with a single contact point if that point is not ideal \cite{cai2024realtosimgrasprethinkinggap}. 

Environments such as construction sites, where large, heavy and irregularly shaped materials are frequently manipulated, exemplify this need. In those contexts, the inherent limitations of single-agent grasping become apparent. For instance, consider the task of lifting a heavy structural beam. A single robotic manipulator, even when executing an optimal grasp, would struggle due to the beam's extended length and weight. These factors increase the likelihood of unstable or failed grasps due to contacts being limited to a single point. Other situations where multi-agency could be leveraged include search and rescue missions (with delicate handling of patients), food preparation, and robotic surgery. A multi-agent controller addresses the issue of requiring close to optimal grasping positions by increasing the number of contact points (which improves grasp stability), and by enabling collaborative behaviors that facilitate smoother and more natural manipulation of complex or irregularly shaped objects.

We propose a multi-agent, collaborative framework of a tactile-reactive Model Predictive Control (MPC) layer. This layer enables coordinated bi-manual manipulation, broadening the range of objects that can effectively be grasped and manipulated, particularly those that are difficult to handle with a single manipulator.

In spite of the many uses of tactile-based controllers, the deployment of these approaches comes with some inherent limitations:
\begin{enumerate}
    \item \textbf{Challenges with soft objects}: Tactile sensors rely on gel deformations to infer qualities about the grasped objects. When interacting with objects that are softer than the sensor's gel, the resulting deformations are often minimal or imperceptible in the captured tactile images. This can lead the controller to interpret such interactions as no-contact scenarios, thus failing to execute appropriate grasping behaviors.
    \item \textbf{Computational Complexity}: Learning-based control methods are naturally computationally expensive, especially when run at high frequencies. They also require a high amount of data points to operate. Homogeneous multi-agent approaches look to make better use of collected experience by sharing parameters across agents. High-frequency control is crucial in manipulation tasks, especially when rapid changes in conditions occur due to collisions or unexpected dynamics, so elevated computing power is required to make these controllers feasible. Two main factors affect controller runtime: the dimensionality of the tactile image encoding and the length of the MPC prediction horizon. Although increasing either of these would theoretically improve prediction accuracy, it also incurs higher computational cost and may lead to optimization issues such as excessively large initial losses during training that yield a non-converging controller. Thus, a balance must be struck between expressiveness and real-time feasibility. 
\end{enumerate}

Our proposed multi-agent tactile-reactive layer addresses the aforementioned challenges as follows:

\begin{enumerate}
    \item \textbf{Bi-manual tactile-reactive grasping of soft objects for improved stability}: By distributing contact forces across two manipulators, our system reduces the likelihood of applying excessive localized pressure. This not only decreases the risk of permanent damage to grasped objects but also provides more stable manipulation through more contact points.
    \item \textbf{A collaborative learning based control approach with improved efficiency}: According to our experimental evaluations, the introduction of an additional manipulator does not result in a proportional increase in runtime due to the leveraging of parameter sharing across the learnable MPC and CNN layers. This indicates that, although the overall system is slower compared to the baseline single-agent case \cite{xuletac2024}, the per-agent efficiency improves. The relative runtime improvement arises mainly from a more efficient construction of the state transition matrices and lifted system matrices.
    \item \textbf{Real-world comparative evaluation between baseline models and our proposed approach}: In our experiments, two GelSight Mini sensors are mounted on Robotiq 2F-series grippers, with all trials conducted under an identical configuration. An extensive evaluation on various objects across multiple baselines is conducted to showcase the advantage of the proposed pipeline.
\end{enumerate}

\begin{figure*}[t]
\minipage{0.16\textwidth} \includegraphics[width=0.99\linewidth]{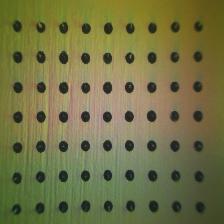} 
\subcaption{Wood} \endminipage\hfill
\minipage{0.16\textwidth} \includegraphics[width=0.99\linewidth]{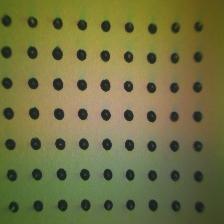} 
\subcaption{Rubber}\endminipage\hfill
\minipage{0.16\textwidth} \includegraphics[width=0.99\linewidth]{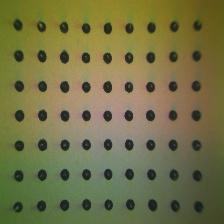} 
\subcaption{Gel}\endminipage\hfill
\minipage{0.16\textwidth} \includegraphics[width=0.99\linewidth]{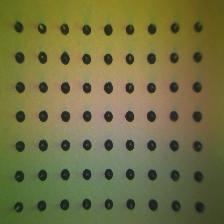} 
\subcaption{Foam} \endminipage\hfill
\minipage{0.16\textwidth} \includegraphics[width=0.99\linewidth]{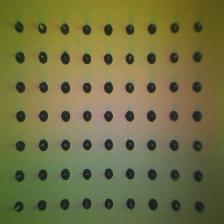} 
\subcaption{Empty}\endminipage\hfill
\caption{Tactile images from materials used during training}
\label{fig:tactile_images}
\end{figure*}

\subsection{Related Works}
\noindent\textbf{Tactile Grasping.}
The main focus of tactile sensors is to mimic the information stream human touch can provide, enabling robotic systems to extract fine-grained information about grasped objects and dynamically adapt their opening and grasping forces. Compared to conventional modalities such as vision or point cloud data, tactile images are significantly richer in encoded contact information, making them especially effective for training data-driven controllers. 
While image-based tactile sensors remain the norm, recent research has explored augmenting them with complementary information streams, such as ultrasound \cite{gong2025ultratacintegratedultrasoundaugmentedvisuotactile} and Vision-Language-Action \cite{huang2025tactilevlaunlockingvisionlanguageactionmodels} \cite{yu2025demonstratingoctopi15visualtactilelanguagemodel}, which have demonstrated promising results in expanding the horizons of tactile sensors. 

A key challenge, however, lies in the lack of tactile sensing datasets and the difficulty of modeling tactile sensors in simulation. Work such as \cite{9981863} has addressed this limitation by leveraging measured contact forces at an end-effector to generate realistic synthetic tactile images, helping to bridge the gap between simulation and deployment. These high-fidelity tactile images could prove very useful in the future, as most current learning-based tactile approaches, including ours, rely exclusively on real sensor data, which can be time-consuming to acquire.

\noindent\textbf{Bi-manual Manipulation.}
In recent years, significant progress has been made in bi-manual robotic manipulation, with many approaches aiming to replicate the dexterity and rich sensing capabilities of human arms. Collaboration between the two manipulators is often achieved through Reinforcement Learning \cite{yuan2024robotsynesthesiainhandmanipulation} or Imitation Learning \cite{grannen2023stabilizeactlearningcoordinate}. A central challenge in this field is accurately discretizing and modeling the environment along with the full range of possible interactions between agents, which makes long-horizon, multi-step tasks particularly complex \cite{zhou2025teachoncelearnoneshot}. This is one of the reasons our approach utilizes parallel-jaw grippers with one degree of freedom: they substantially reduce the action space compared to multi-finger grippers, simplifying the learning process and making it more tractable.

\noindent\textbf{Learning-based MPC.}
Model Predictive Control (MPC) is one of the most robust and widely accepted control frameworks in robotics. However, traditional MPC approaches often face limitations when applied to systems that are inherently difficult to model with high fidelity. Learning-based MPC seeks to overcome these challenges by incorporating real-world (sensory) data to refine system dynamics, enabling the design of more accurate and adaptive controllers. In many cases, learning-based MPC has been shown to achieve superior performance, streamline deployment, and reduce the need for extensive manual controller tuning compared to baseline MPC methods \cite{annurev:/content/journals/10.1146/annurev-control-090419-075625}. Applications of MPC span a wide range of domains, including dexterous manipulation \cite{todorov2016manipulation}, autonomous vehicle steering \cite{zhang2024applicationdatadrivenmodelpredictive}, and Unmanned Aerial Vehicles (UAVs) \cite{bui2024modelpredictivecontroloptimal}.

\subsection{Organization of the paper}
The remainder of the paper is organized as follows: In the \textit{Data Collection} section, we provide an overview of the methods we employed to obtain data for training our learned controller.
In the \textit{Pipeline Architecture} section, we describe the individual components of the proposed system and their interactions, outlining how they collectively enable multi-agent tactile reactive grasping. In the \textit{Results} section, we report on training and experiment results, comparing the performance of our method against established baselines. Following that, in the \textit{Advantages over Single-agent Model} section, we present a comparative analysis highlighting the benefits of the proposed layer relative to baseline single-agent approaches. Lastly, in the \textit{Conclusion, Limitations and Future Work} section, we summarize the contributions, discuss the current limitations, and outline potential directions for future research.

\section{Data Collection for Bi-Manual Tactile-Grasping}

Our data collection pipeline is multi-agent in nature, leveraging two Robotiq 2f-series grippers, each equipped with a Gelsight Mini tactile sensor near the fingertips of the end effector. This setup enables simultaneous data acquisition for both arms during contact-rich manipulation.

\begin{figure}[h!]
    \centering
    \begin{subfigure}{0.35\linewidth}
        \centering
        \includegraphics[width=\linewidth]{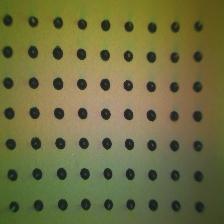}
        \caption{Heavy contact}
        \label{fig:heavy_contact}
    \end{subfigure}
    \begin{subfigure}{0.35\linewidth}
        \centering
        \includegraphics[width=\linewidth]{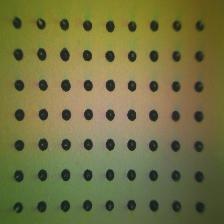}
        \caption{Partial slip}
        \label{fig:pslip}
    \end{subfigure}
    \caption{Tactile sensor outputs under different conditions.}
    \label{fig:tactile_conditions}
\end{figure}

The objective of the data collection process is to capture tactile images across a range of contact conditions between the sensors and the surface of standardized blocks. In our approach, we utilized wooden, soft gel, soft foam, and hard rubber surfaces. During the trials, we aim to identify and record the transition between firm contact and tactile slippage. To this end, we define the \textit{slippage opening} as the minimum gripper width (in millimeters) at which an object begins to slip due to insufficient frictional force between the gel membrane and the object's surface. As evidenced in Figure~\ref{fig:tactile_conditions}, our dataset includes tactile image sequences spanning from full contact to this critical point of slippage.
Our approach involves manual measurement of two key parameters: the initial contact width and the slippage threshold.  To enhance dataset diversity and improve generalization, we introduce small random variance to both the initial and slippage openings across trials. Specifically, the initial width is varied by $\pm 0.7\,\text{mm}$, and the \textit{slippage opening} is varied by $\pm 0.35\,\text{mm}$.

In addition to the four standardized material blocks used for contact trials, we also collect no-contact baseline images, representing scenarios in which neither gripper is in contact with an object. These samples are essential for training the model to recognize and respond to surface contact changes. During no-contact situations, both grippers are commanded to gradually close until they make contact with an object.

\begin{figure*}[t]
    \centering
    \begin{subfigure}{0.55\linewidth}
        \includegraphics[width=\linewidth]{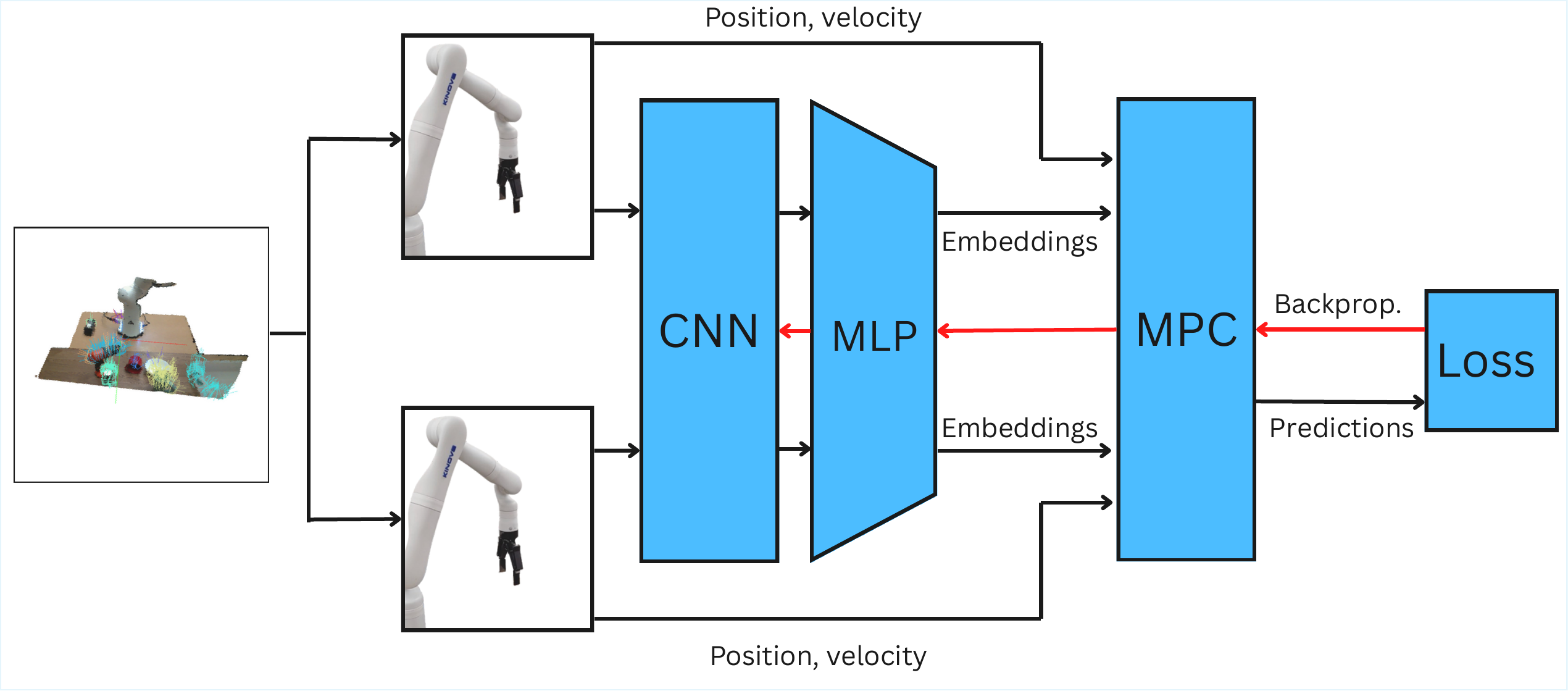}
        \caption{Pipeline architecture.}
        \label{fig:pipeline}
    \end{subfigure}\hfill
    \begin{subfigure}{0.45\linewidth}
        \includegraphics[width=\linewidth]{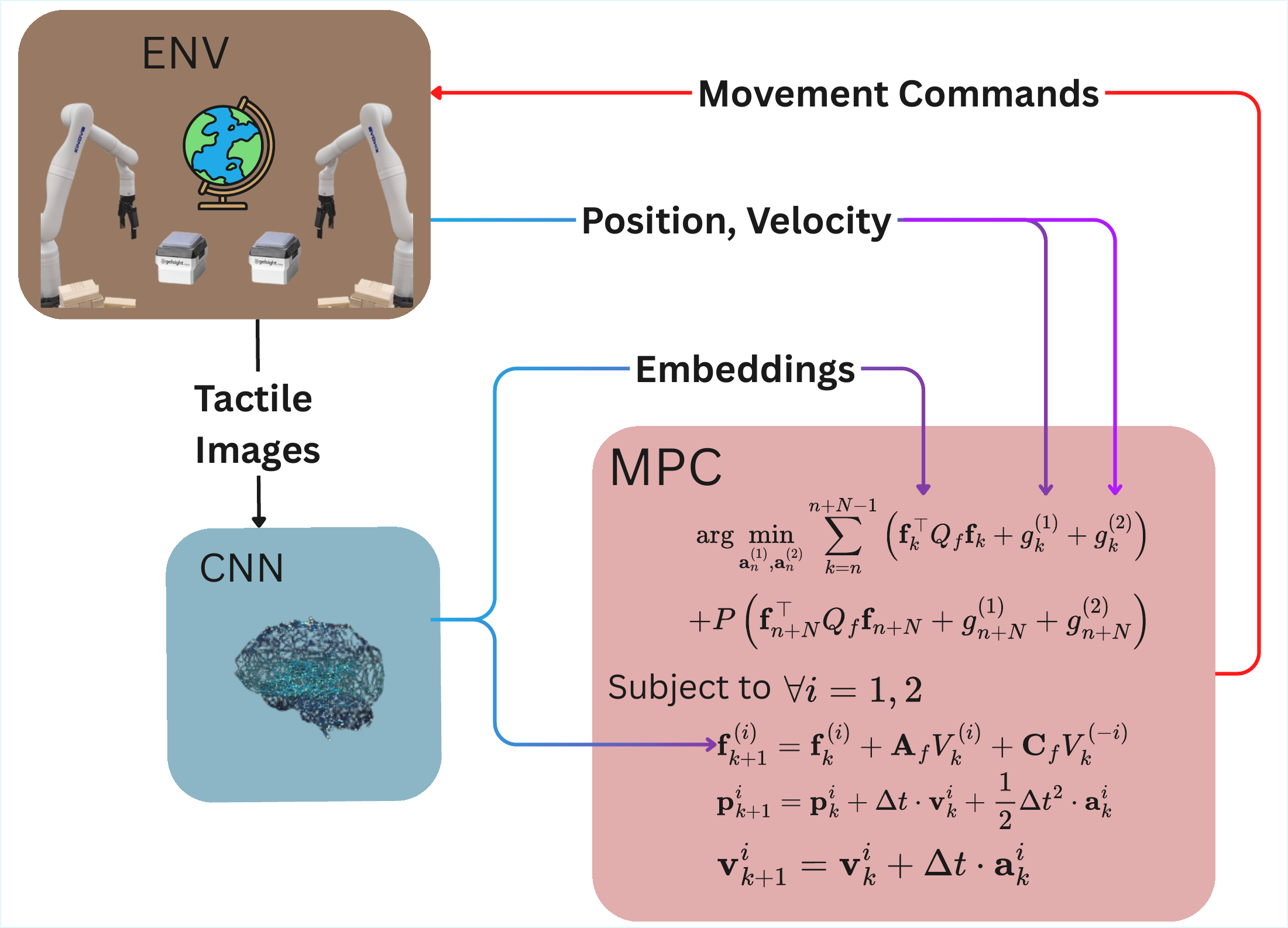}
        \caption{MPC layer.}
        \label{fig:mpc}
    \end{subfigure}
    \caption{Architecture of our collaborative learning based multi-agent MPC with tactile sensory input.}
    \label{fig:combined}
\end{figure*}

Each data collection trial consists of four sub-trials, structured as follows:
\begin{enumerate}
    \item Both grippers begin by grasping the target object.
    \item One gripper (designated as the "active" gripper) gradually opens while the other remains stationary. This continues until the active gripper reaches the predefined \textit{slippage opening}. Tactile images are recorded continuously from both sensors throughout this motion.
    \item Upon reaching the slippage threshold, the active gripper returns to its initial position, the stationary gripper is slightly opened, and the process is repeated with roles reversed. This continues until both grippers have independently reached the \textit{slippage opening}.
\end{enumerate}
Following each sub-trial, 25 pairs of tactile images are saved in a directory labeled with the trial number, sub-trial index, and measured \textit{slippage opening} (in millimeters). Each image filename is prefixed with "1" or "2" to indicate what sensor it belongs to, and filenames include the current gripper opening and a frame index to enable precise alignment during model training.
To ensure balanced data representation across both agents, the roles of "active" and "stationary" gripper are alternated between trials.

\section{Collaborative Learning-Based Multi-Agent MPC with Tactile Sensory Input}

We adopt a modified ResNet-152 architecture as our tactile image encoder. We freeze the convolutional backbone to preserve pre-trained visual features and replace the original classification head with a two-layer multilayer perceptron (MLP) with ReLU activations. The MLP compresses high-dimensional tactile images into low-dimensional embeddings $f \in \mathbb{R}^M$, which are used as inputs to our MPC module. The hidden layers of the MLP have a dimensionality of 128, and the final output embedding size is set to $ \mathbb{M} = 20$.

For each forward pass, the MPC layer operates on a pair of embeddings; one per agent, as well as the current position and velocity values for both agents. To ensure consistent interpretation of these inputs, it is critical to maintain a fixed agent ordering throughout both training and inference. This requirement stems from the inherent variability among Gelsight sensors: even when subjected to identical contact conditions, two different sensors produce slightly distinct tactile embeddings due to subtle manufacturing and calibration differences. As such, any mismatch in input ordering has the potential to negatively impact model performance.

For position and velocity, we use metric units (millimeters and millimeters per second, respectively). Although the Robotiq gripper provides normalized position readings in the range [0.0, 0.8], we convert to metric units to improve compatibility with other robotic platforms. This design choice enables greater generalizability, allowing the learned policy to be deployed on alternative hardware platforms, provided they can report opening width and velocity in metric units, which is a reasonable assumption in the current robotic landscape.

\subsection{Layer of Multi-agent Learning-Based MPC }
We operate on the assumption that the agents have one degree of freedom (open or close along a singular axis). Then, denoting the sampling time interval at $\Delta t$, the gripper motion is defined using a discrete-time double integrator model as: 
\begin{equation}
\begin{bmatrix}
\mathbf{p^{(i)}_{n+1}} \\
\mathbf{v^{(i)}_{n+1}}
\end{bmatrix}
=
A_g
\begin{bmatrix}
\mathbf{p^{(i)}_n} \\
\mathbf{v^{(i)}_n}
\end{bmatrix}
+
B_g \mathbf{a^{(i)}_n}, \forall i=1,2
\end{equation}
where
\[
A_g =
\begin{bmatrix}
1 & \Delta t \\
0 & 1
\end{bmatrix}
\in \mathbb{R}^{2 \times 2}, \quad
B_g =
\begin{bmatrix}
\frac{1}{2} \Delta t^2 \\
\Delta t
\end{bmatrix}
\in \mathbb{R}^2, \\
\]
$\mathbf{p}, \mathbf{v}, \mathbf{a}$ are scalar variables representing the position (in $mm$), velocity (in $mm/s$) and acceleration (in $mm/s^2$), respectively. The superscript $^{(i)}$ denotes the agent index.
The main optimization problem in the MPC layer is the following (where $\mathbf{f}_k$ is the tactile embeddings hidden state, \bm{$p_k$} and \bm{$v_k$} are the position and velocity at timestep $k$, and $P$ amplifies the terminal cost weight): 
\begin{align}
\begin{split}
\mathbf{a}_n^{(1*)}, \mathbf{a}_n^{(2*)} &= \arg \min_{\mathbf{a}_n^{(1)}, \mathbf{a}_n^{(2)}}  \sum_{k=n}^{n+N-1} \left( \mathbf{f}_k^\top Q_f \mathbf{f}_k + g^{(1)}_k + g^{(2)}_k \right) \\
&\quad  \ + P\left(\mathbf{f}_{n+N}^\top Q_f \mathbf{f}_{n+N} + g^{(1)}_{n+N} + g^{(2)}_{n+N} \right)
\label{eq:cost_function}
\end{split} 
\end{align}
\text{with:} 
\begin{align}
\mathbf{f}_k &\triangleq
\begin{bmatrix}
\mathbf{f}^{(1)}_k \\
\mathbf{f}^{(2)}_k
\end{bmatrix}
\label{eq:f_definition}
 \\
g^{(i)}_k &\triangleq Q_v (\mathbf{v}^{(i)}_k)^2 +Q_a(\mathbf{a}_k^{(i)})^2, \forall i=1,2
\end{align}

\text{Subject to:} $\forall i=1,2$
\begin{align}
\mathbf{f}_{k+1}^{(i)} &= \mathbf{f}_{k}^{(i)} + \mathbf{A}_{f}{V}_{k}^{(i)} + \mathbf{C}_{f}{V}_{k}^{(-i)}  
\label{eq:f_transition}\\
\mathbf{p}_{k+1}^{(i)} &= \mathbf{p}_{k}^{(i)} + \Delta t \cdot \mathbf{v}_k^{(i)} + \frac{1}{2} \Delta t^2 \cdot \mathbf{a}_k^{(i)}
\label{eq:p_transition}\\
\mathbf{v}_{k+1}^{(i)} &= \mathbf{v}_k^{(i)} + \Delta t \cdot \mathbf{a}_k^{(i)}
\label{eq:v_transition}
\end{align}
where $\mathbf{f}_k^\top Q_f \mathbf{f}_k$ is the learnable cost for tactile sensory input, 
$Q_a(\mathbf{a}_k^{(i)})^2$ penalizes the magnitude of the actual action to prevent excessive control effort. Note that $\mathbf{f}_k^\top Q_f \mathbf{f}_k$ and $\mathbf{C}_{f}$ capture the coupling between two grippers/ tactile sensors, which reflects how the tactile state of an agent influences and relates to that of the other agent and encourages cooperative behavior. Specifically, the former characterizes the coupling in the loss function as  $\mathbf{f}_k$ is the concatenation of the sensory input from both agents, while the latter directly models the coupling in dynamics.

The proposed MPC layer is fully differentiable, enabling it to be seamlessly integrated onto learning pipelines along conventional neural network components. We utilize OSQP \cite{stellato2020osqp}, a numerical optimization solver designed specifically for convex Quadratic Problems (QPs) of the form:
\begin{align}
\min_{x} \quad & \frac{1}{2} x^\top P x + q^\top x \\
\text{subject to} \quad & l \leq A x \leq u \\
& 
\begin{bmatrix}
p_{\min} \\
v_{\min} \\
a_{\min}
\end{bmatrix}
\le
\begin{bmatrix}
p_{n} \\
v_{n} \\
a_{n}
\end{bmatrix}
\le
\begin{bmatrix}
p_{\max} \\
v_{\max} \\
a_{\max}
\end{bmatrix}
\end{align}
Where the decision vector $x$ encodes the predicted control actions, the cost matrices $P$, $q$, and constrains $A$, $l$, $u$ depend on: 
\begin{itemize}
    \item The tactile embeddings $f_k$
    \item The system dynamics ($A_g, B_g$)
    \item The weight matrix ($Q_f$)
\end{itemize}
Similar to layers in a deep network, our MPC layer takes inputs, solves a structured optimization problem, and allows gradients to backpropagate through the solution. In addition, this layer is fully end-to-end differentiable and trainable, enabling joint optimization of the controller parameters alongside other components of the network. This differentiability is crucial for learning based control: it enables the controller to adjust its initial parameters through standard gradient-based optimization \cite{ruder2017overviewgradientdescentoptimization}. 

\subsection{Training}
The overall loss function used during training is: \\
\begin{align}
\mathcal{L} 
&= \text{MSE}( \hat{Y}_{\text{1}}, Y_{\text{1}} ) 
+ \text{MSE}( \hat{Y}_{\text{2}}, Y_{\text{2}} ) \\
&\quad + \text{MSE}( \,S \cdot \hat{y}^{(T)}_{\text{1}}, \,S \cdot y^{(T)}_{\text{1}} ) 
+ \text{MSE}( \,S \cdot \hat{y}^{(T)}_{\text{2}}, \,S \cdot y^{(T)}_{\text{2}} ) \nonumber
\end{align}
{where:}
\begin{align*}
\hat{Y}_{\text{1}}, \hat{Y}_{\text{2}} & : \text{ predicted trajectories (all timesteps)} \\
Y_{\text{1}}, Y_{\text{2}} & : \text{ ground-truth trajectories (expanded)} \\
\hat{y}^{(T)}_{\text{1}}, \hat{y}^{(T)}_{\text{2}} & : \text{ last-timestep predictions} \\
y^{(T)}_{\text{1}}, y^{(T)}_{\text{2}} & : \text{ last-timestep ground truths} \\
\text{S}  & : \text{ Scaling for the last timestep}
\end{align*}

The metadata saved during data collection, corresponding to the beginning of slippage, is used to determine the values for $y^{(T)}_{\text{1}} \text{and } y^{(T)}_{\text{2}}$. For each trial, the model receives information depicting openings ranging from the initial state to the slippage point, and it is expected to generate outputs that closely approximate the true values specified in the metadata for the slippage opening.
We introduce a scalar weight $S = 3$ to emphasize the importance of the final timestep in the prediction horizon, thereby assigning greater significance to whether the controller ultimately reaches the desired position. 
Empirically, $S = 3$ was found to be most effective: lower values provided insufficient emphasis, while higher values led to excessively large losses, affecting the convergence ability for our model.
The parameters used for the MPC layer during training are found in Table I, and the parameters used for the CNN encoder during training are found in Table II. 

Let \textit{M} denote the dimensionality of the tactile embeddings produced by our ResNet-152 CNN encoder, which takes tactile images as inputs. Let \textit{N} denote the prediction horizon on which the MPC operates. We train the controller end-to-end by optimizing the following learnable parameters: 
\begin{enumerate}
    \item \textbf{ $\mathbf{A_f\in\mathbb{R}^{N\times 1}}$:} A component of the primary state transition matrix, representing how an agent's velocity influences its own hidden state over the prediction horizon. Present in Equation \eqref{eq:f_transition}.
    \item \textbf{ $\mathbf{C_f\in\mathbb{R}^{1\times N}}$:} a coupling term that, together with $\mathbf{A_f}$, forms the full state transition matrix. Positioned in the off-diagonal blocks, $\mathbf{C_f}$ encodes how the velocity of one agent affects the hidden state of the other. Present in Equation \eqref{eq:f_transition}.
    \item \textbf{ $\mathbf{Q_f\in\mathbb{R}^{2(N+2)\times 2(N+2)}}$:}  Learnable quadratic penalty matrix of the tactile state embeddings $\mathbf{f}$, considering the coupling of two agents. Present in Equation \eqref{eq:cost_function}.
    \item $\mathbf{\alpha}$: a learned scalar coefficient used to scale the coupling blocks within the $Q$ matrix. Without appropriate scaling, the off-diagonal coupling can dominate, resulting in a loss of positive semi-definiteness (PSD) in the cost matrix.
\end{enumerate}

\begin{figure}[t]
\minipage{0.08\textwidth} \includegraphics[rotate=-90, width=0.99\linewidth]{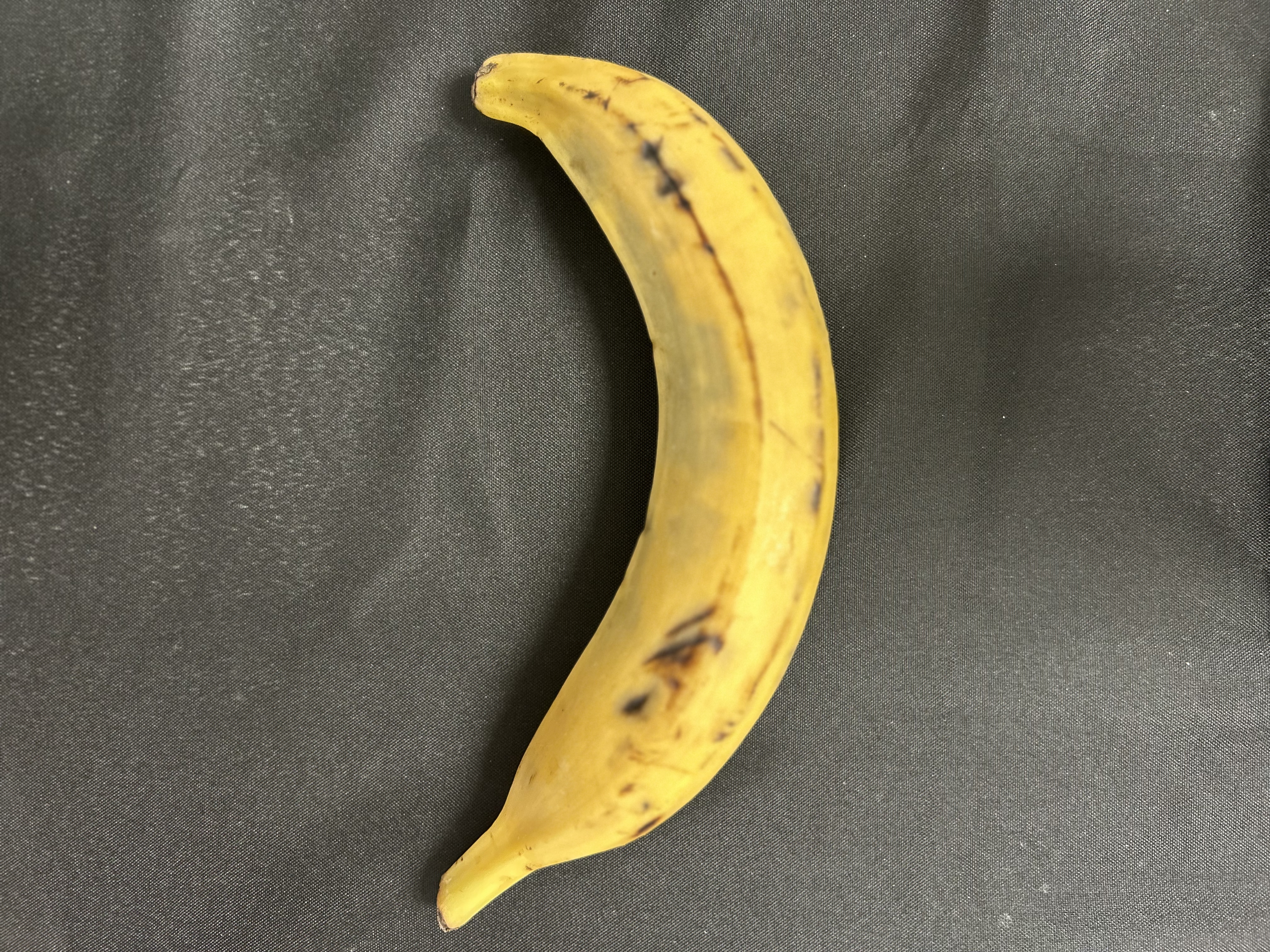} 
\endminipage\hfill
\minipage{0.08\textwidth} \includegraphics[width=0.99\linewidth]{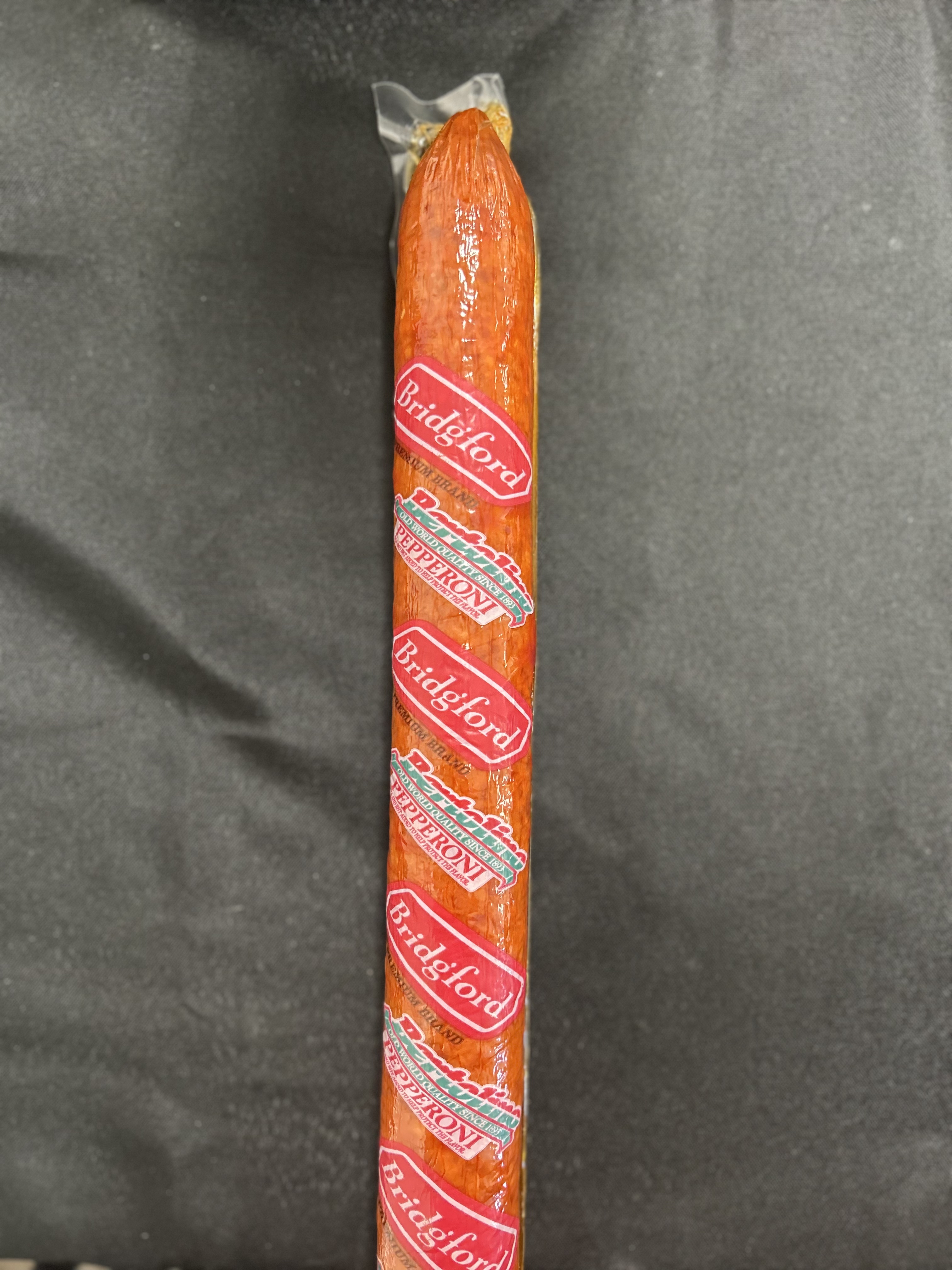} 
\endminipage\hfill
\minipage{0.08\textwidth} \includegraphics[width=0.99\linewidth]{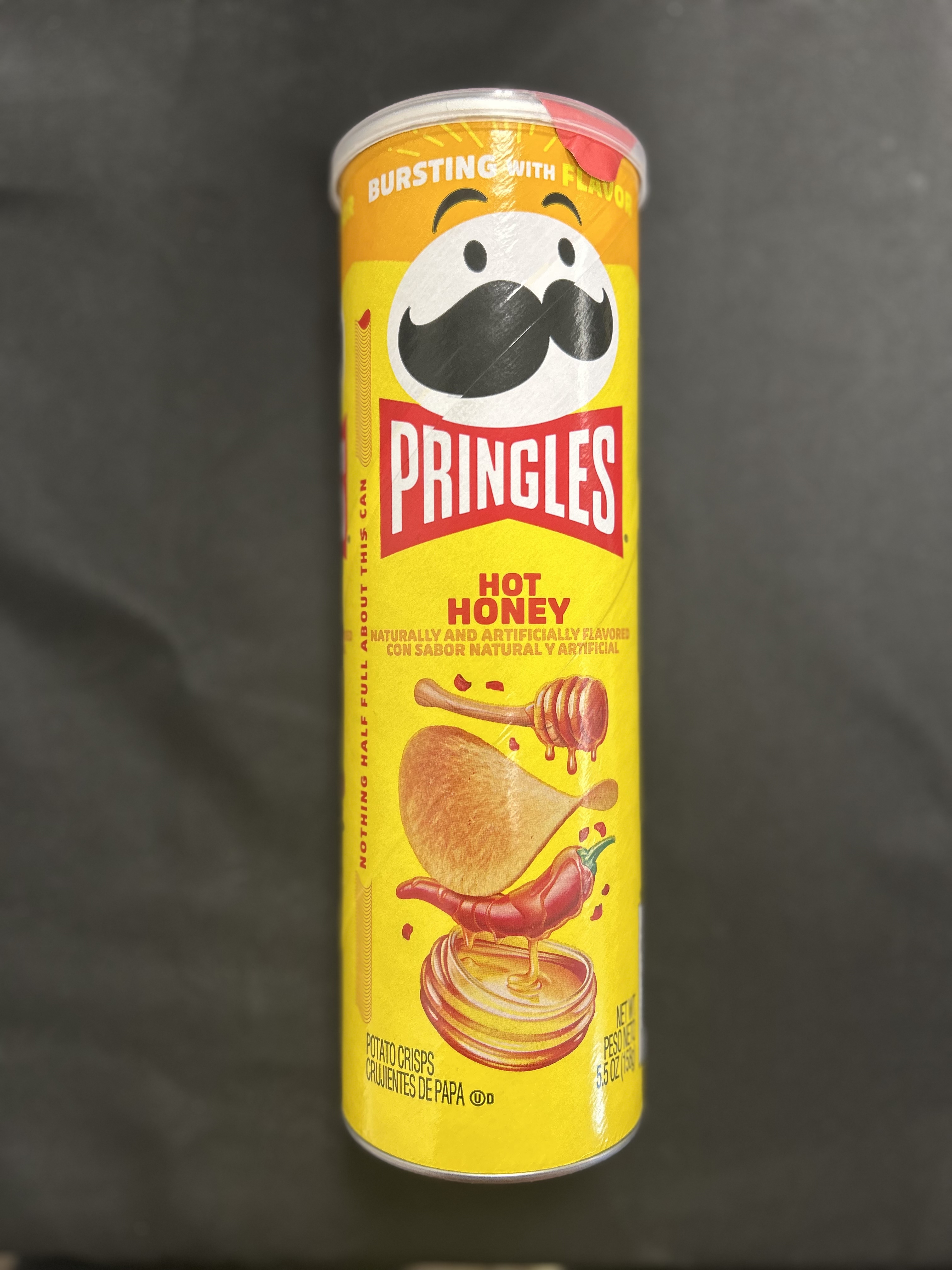} 
\endminipage\hfill
\minipage{0.08\textwidth} \includegraphics[angle=-90, width=0.99\linewidth]{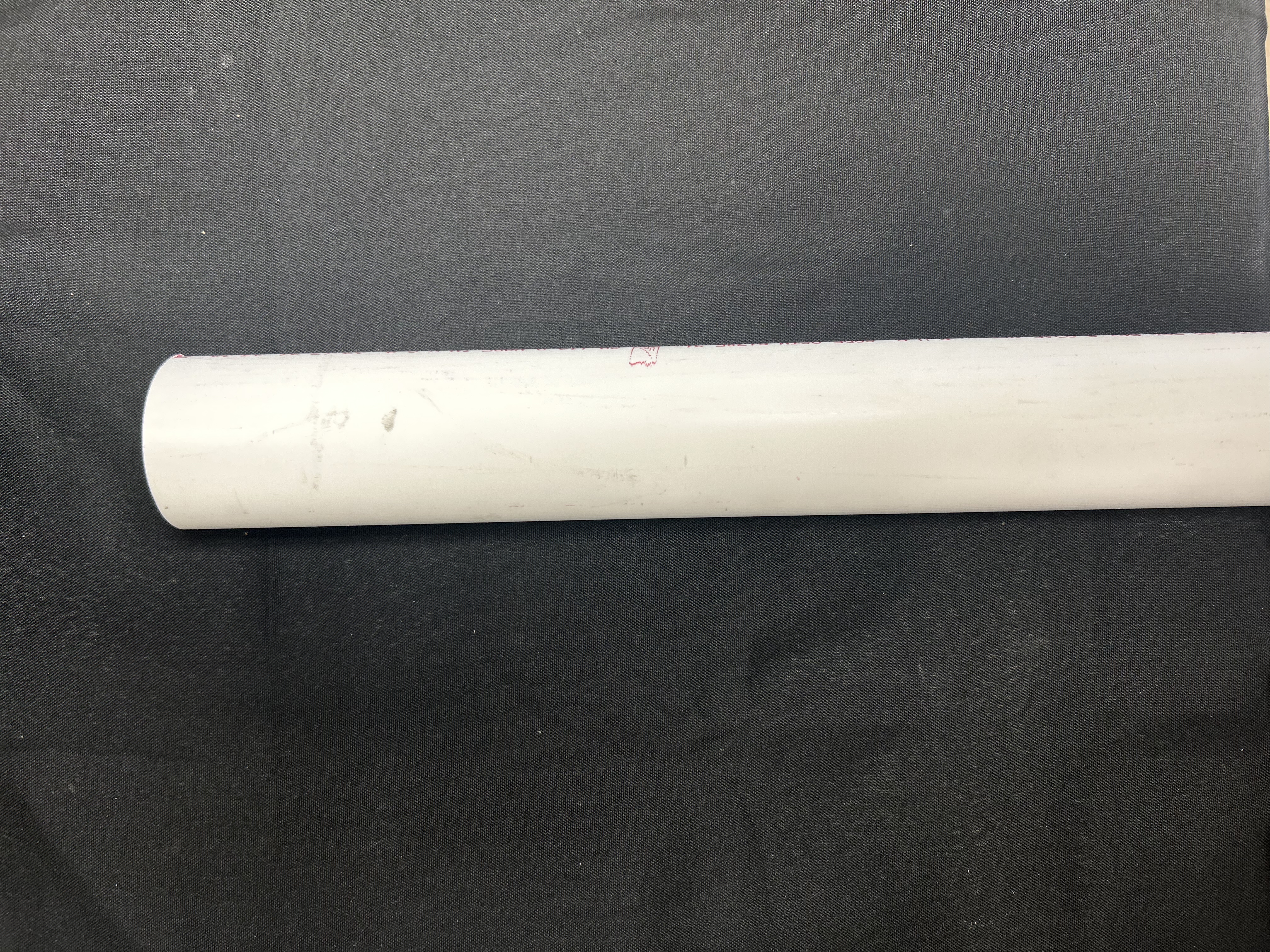} 
\endminipage\hfill
\minipage{0.08\textwidth} \includegraphics[width=0.99\linewidth]{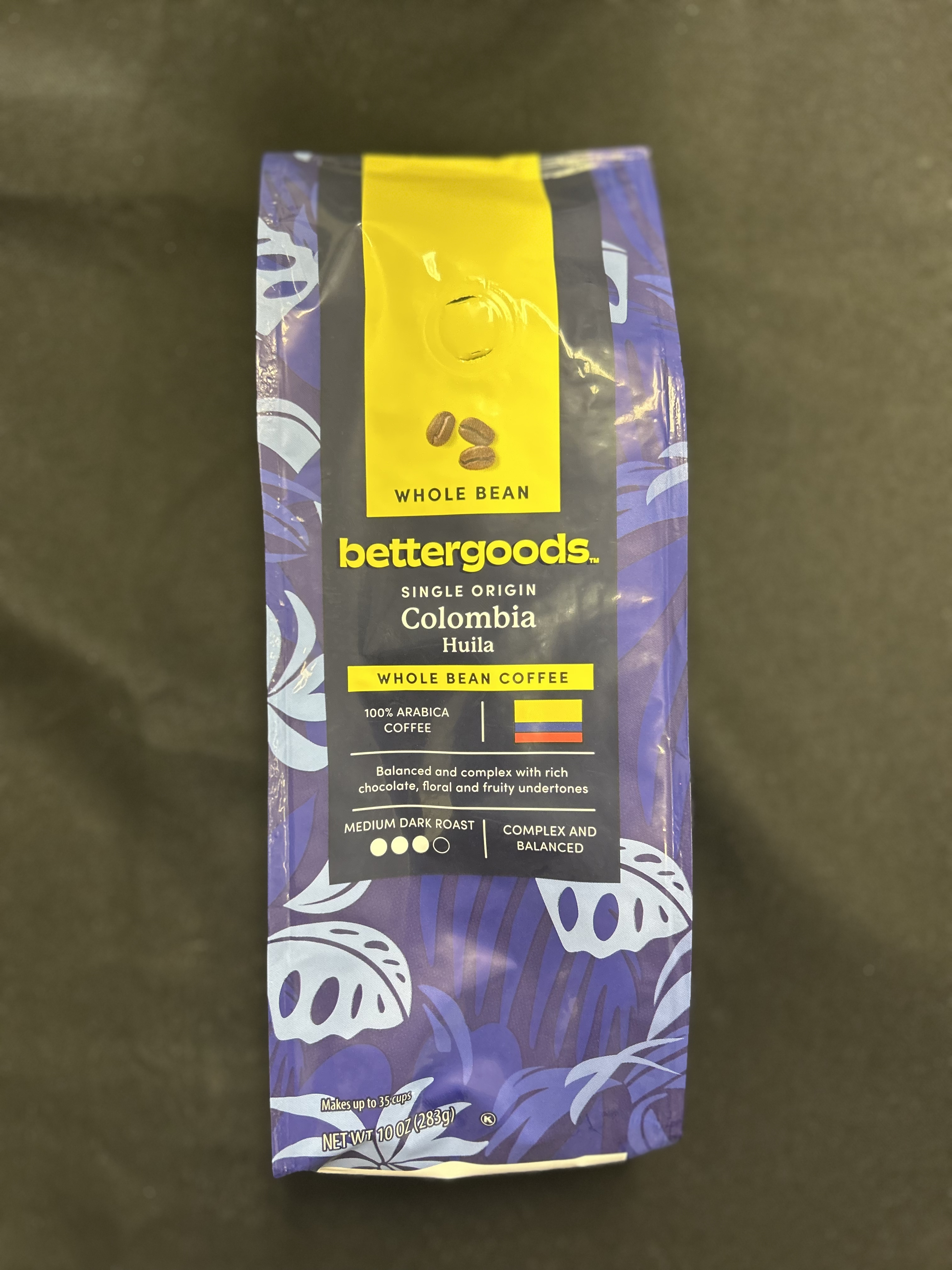} 
\endminipage\hfill
\caption{Materials used for real-world experiments: from left to right: Banana, Pepperoni, Pringles Can, PVC Pipe, and Coffee Beans Bag.}
\label{fig:trial_objects}
\end{figure}

\begin{figure*}[t]
\centering
\begingroup
\newcommand{\InclGr}[1]{\includegraphics[width=0.31\textwidth,valign=t]{#1}}%
\begin{tabular}{c c c}
\hline\hline
\multicolumn{1}{c}{Multi-agent MPC (\textbf{Ours})} & 
\multicolumn{1}{c}{Single-agent MPC} & 
\multicolumn{1}{c}{Single-agent PD} \\
\hline
\multicolumn{3}{c}{Banana} \\
\InclGr{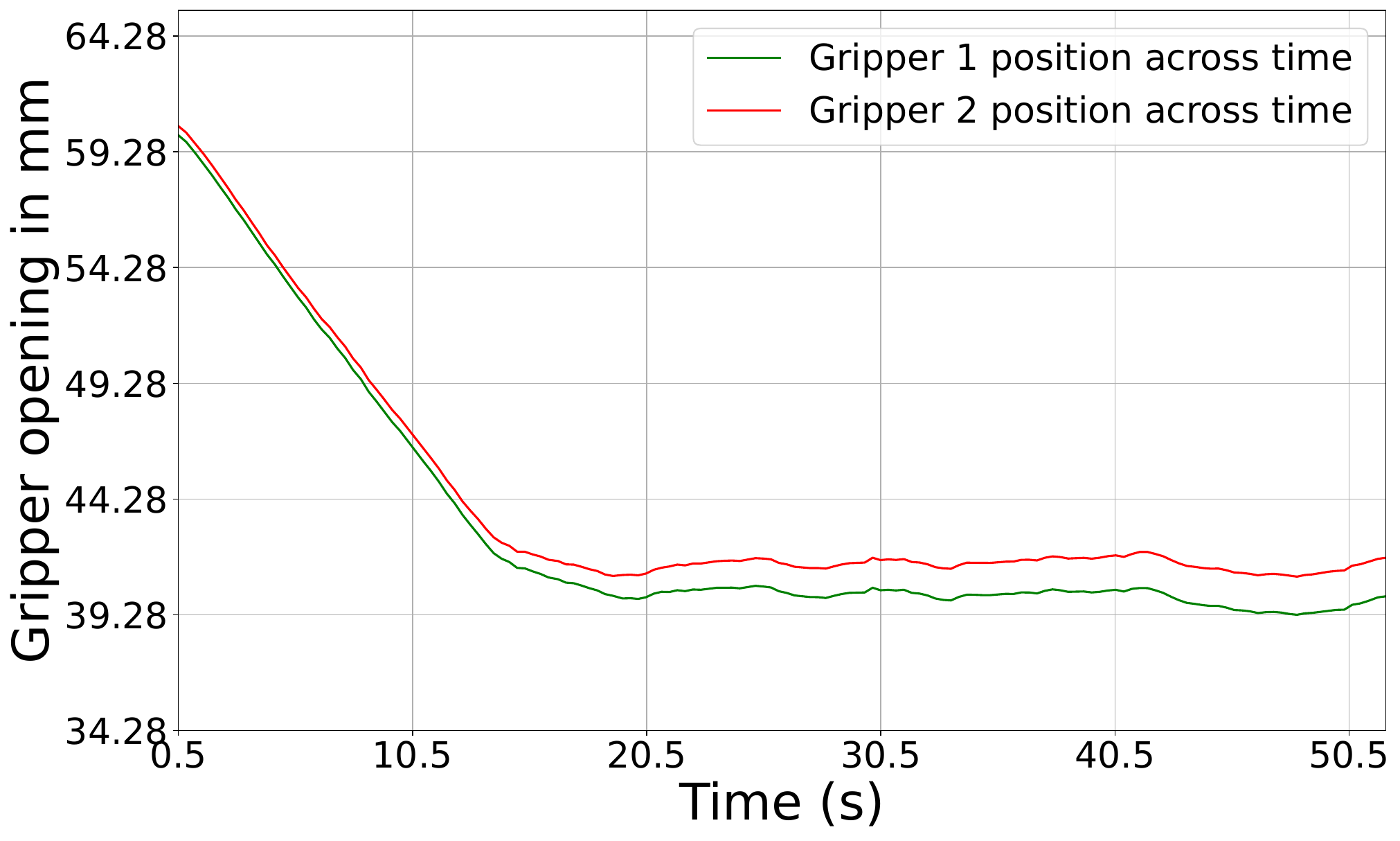} & 
\InclGr{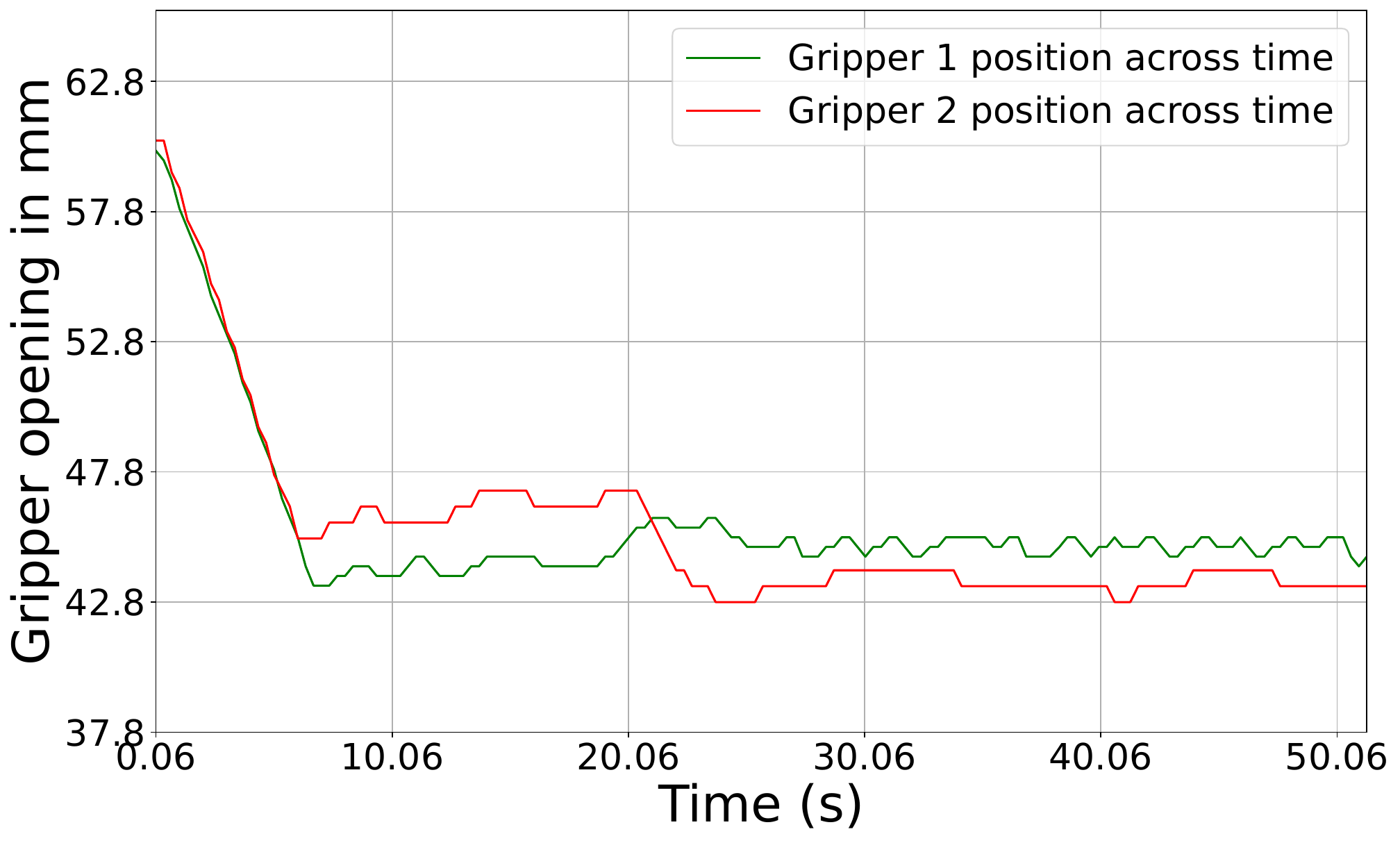} & 
\InclGr{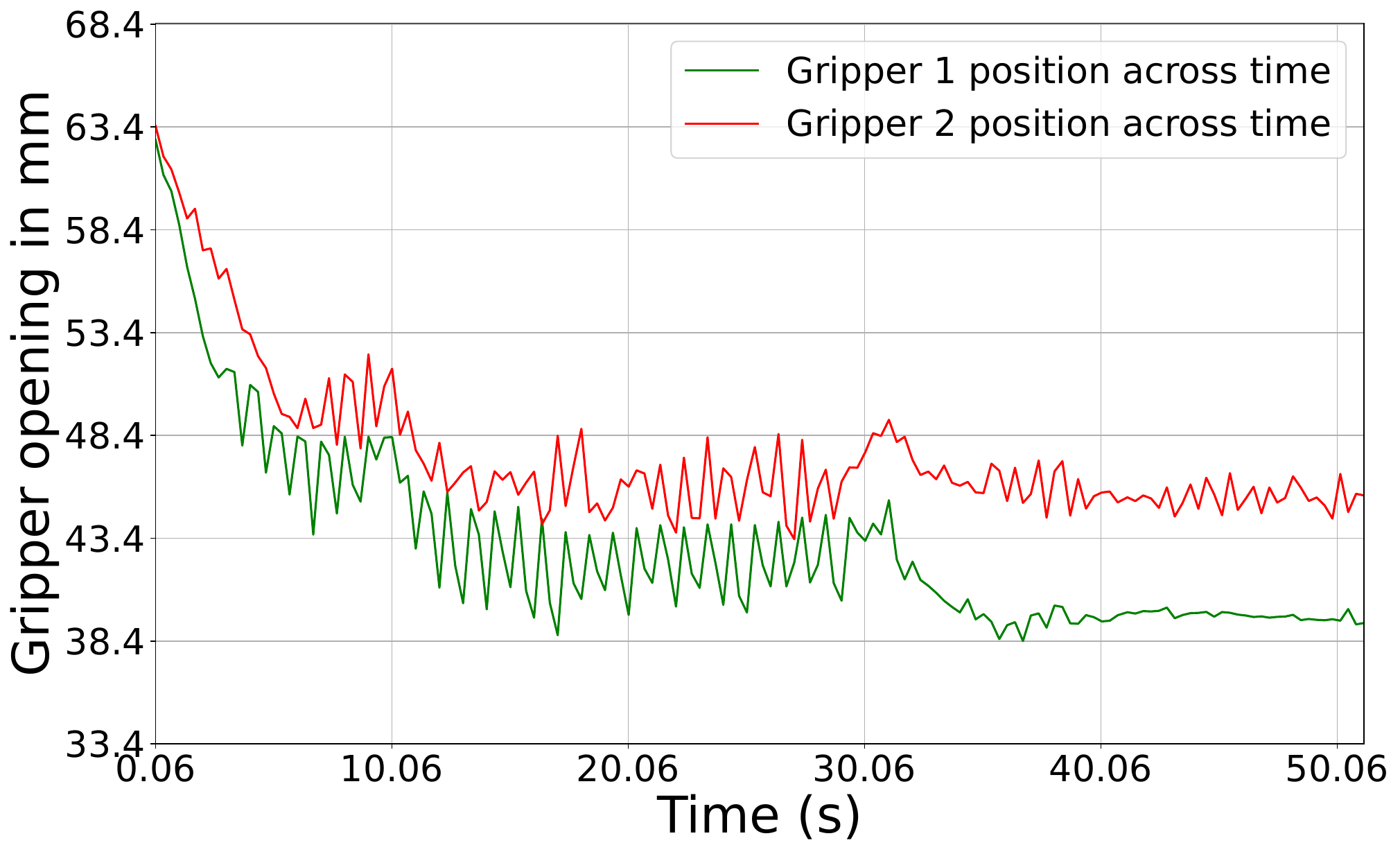} \\
\hline
\multicolumn{3}{c}{Pepperoni Stick} \\
\InclGr{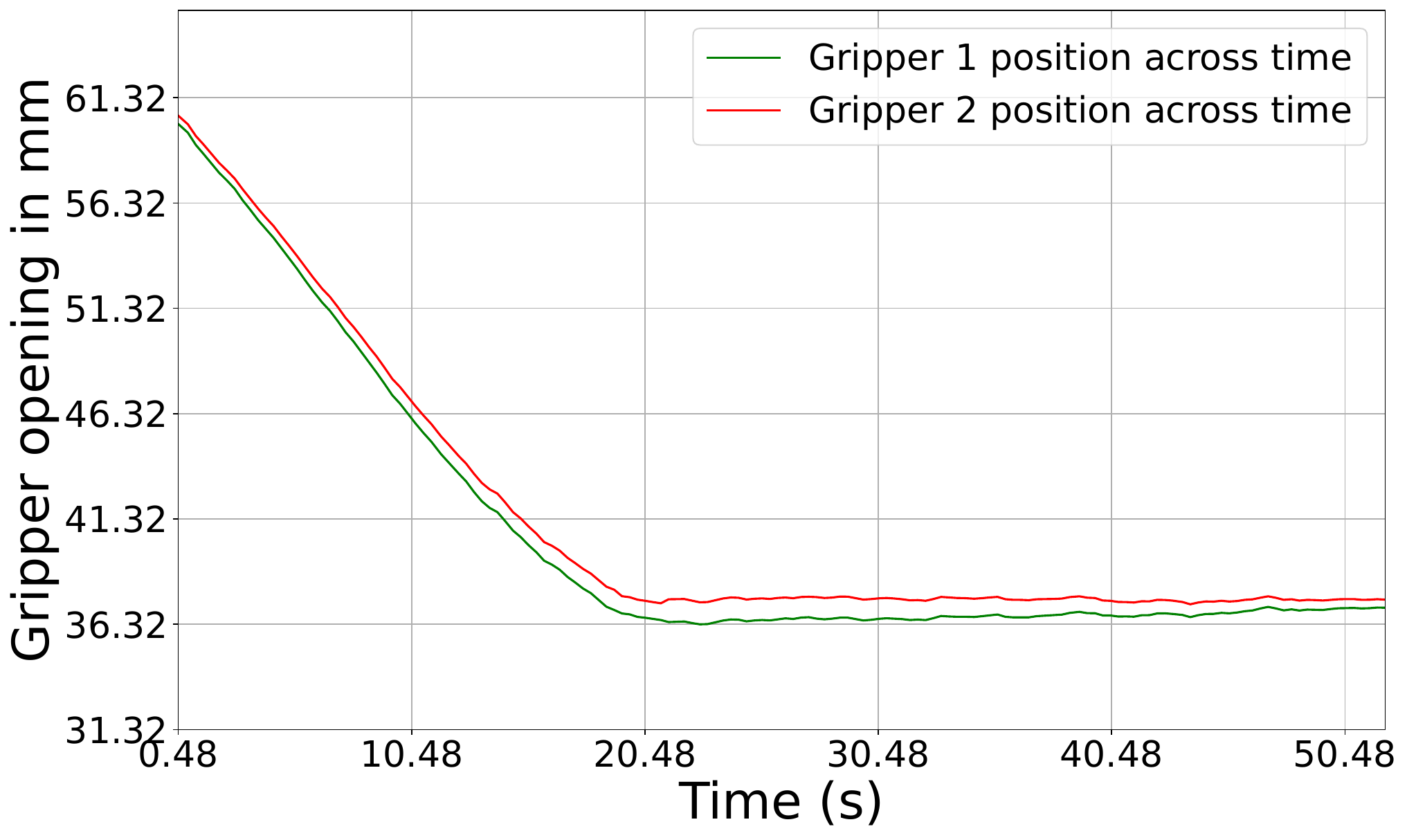} & 
\InclGr{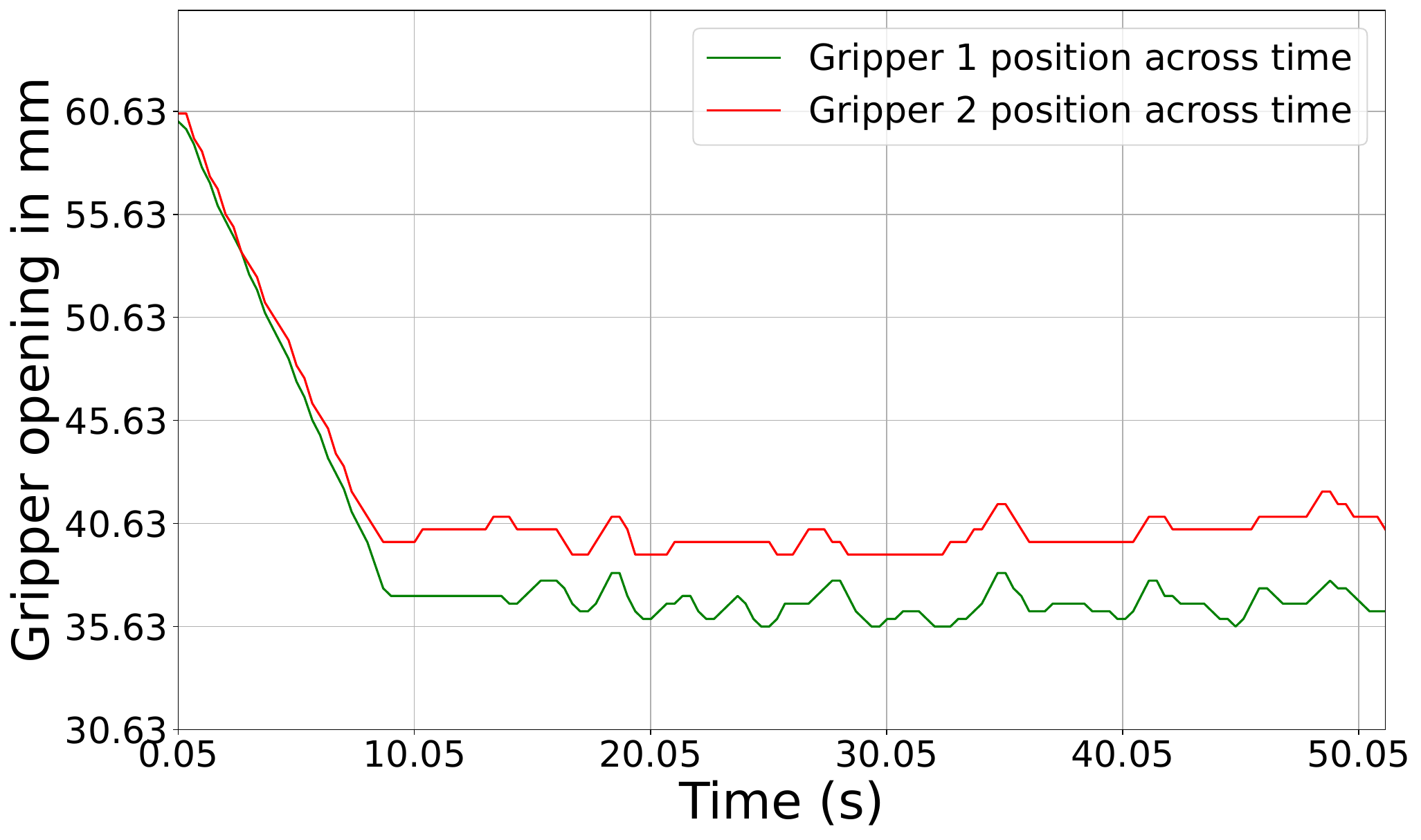} & 
\InclGr{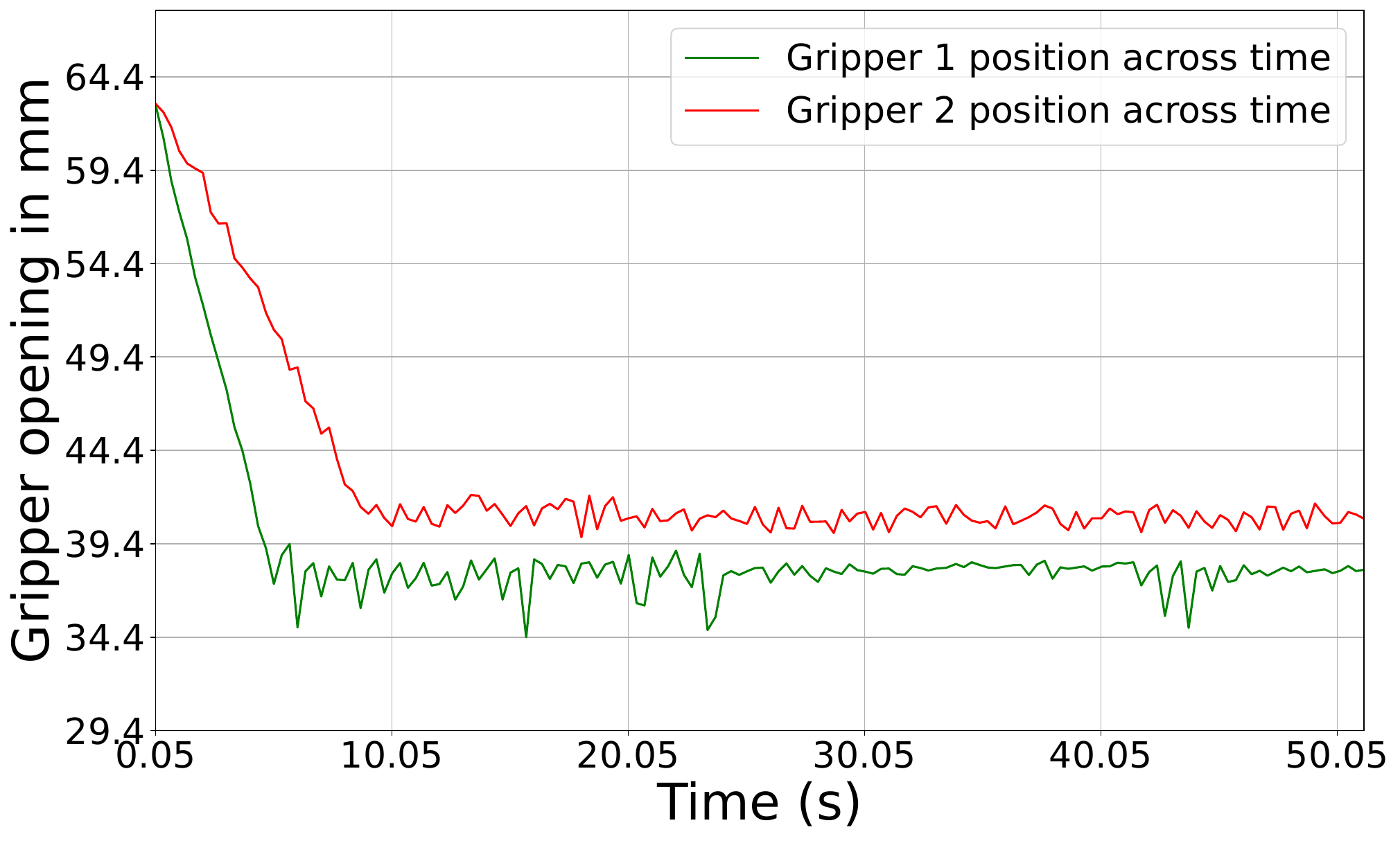} \\
\hline
\multicolumn{3}{c}{Pringles Can} \\
\InclGr{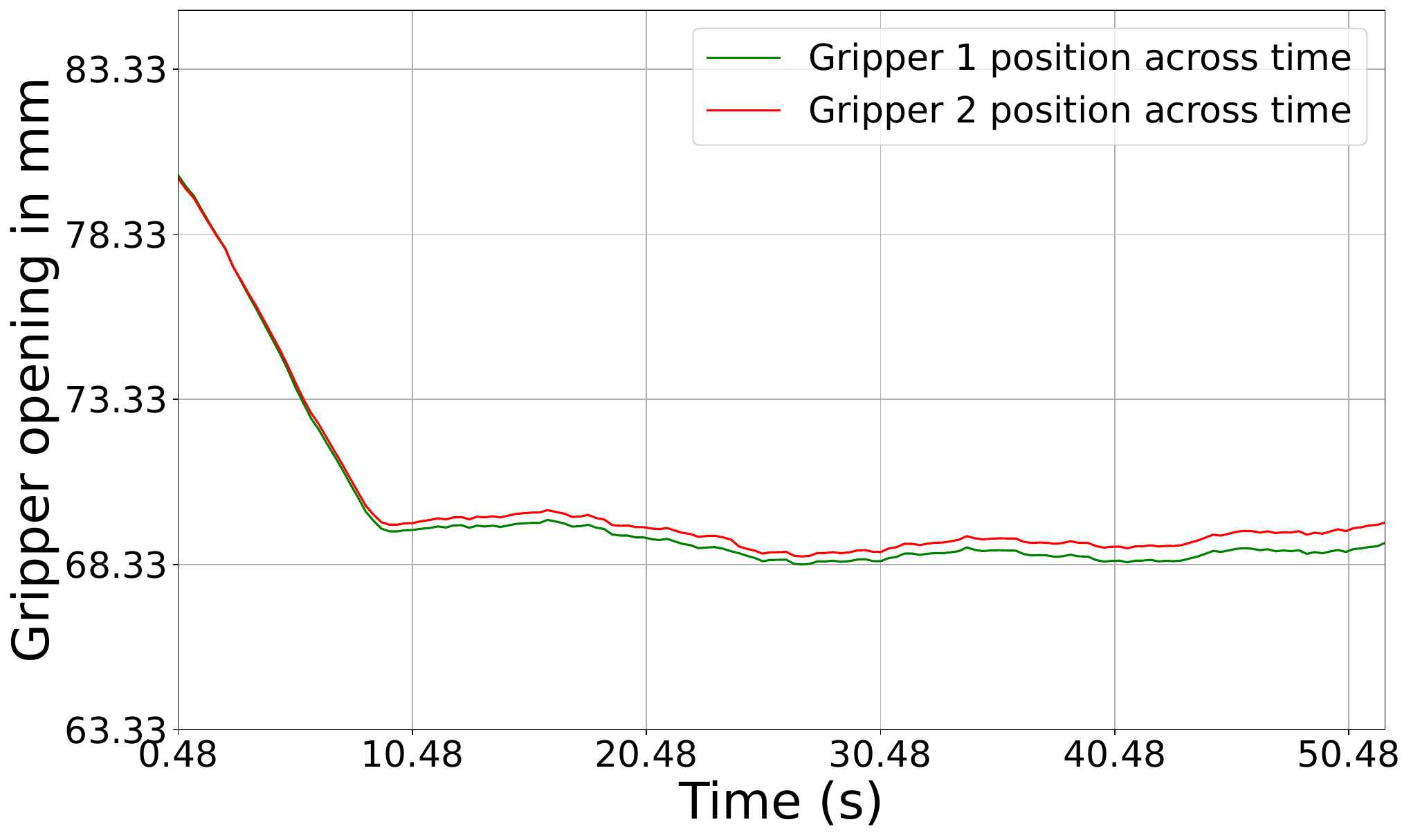} & 
\InclGr{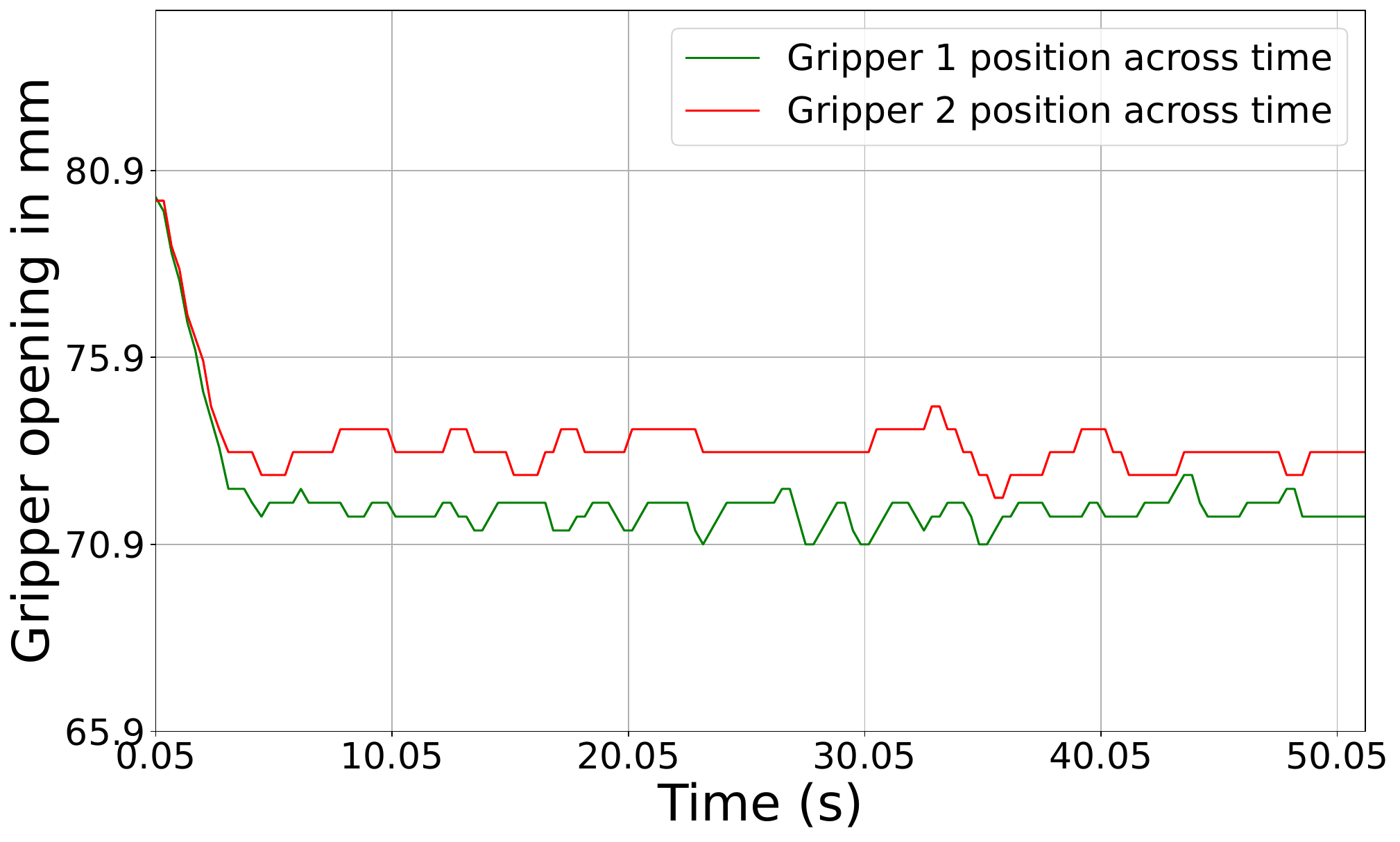} & 
\InclGr{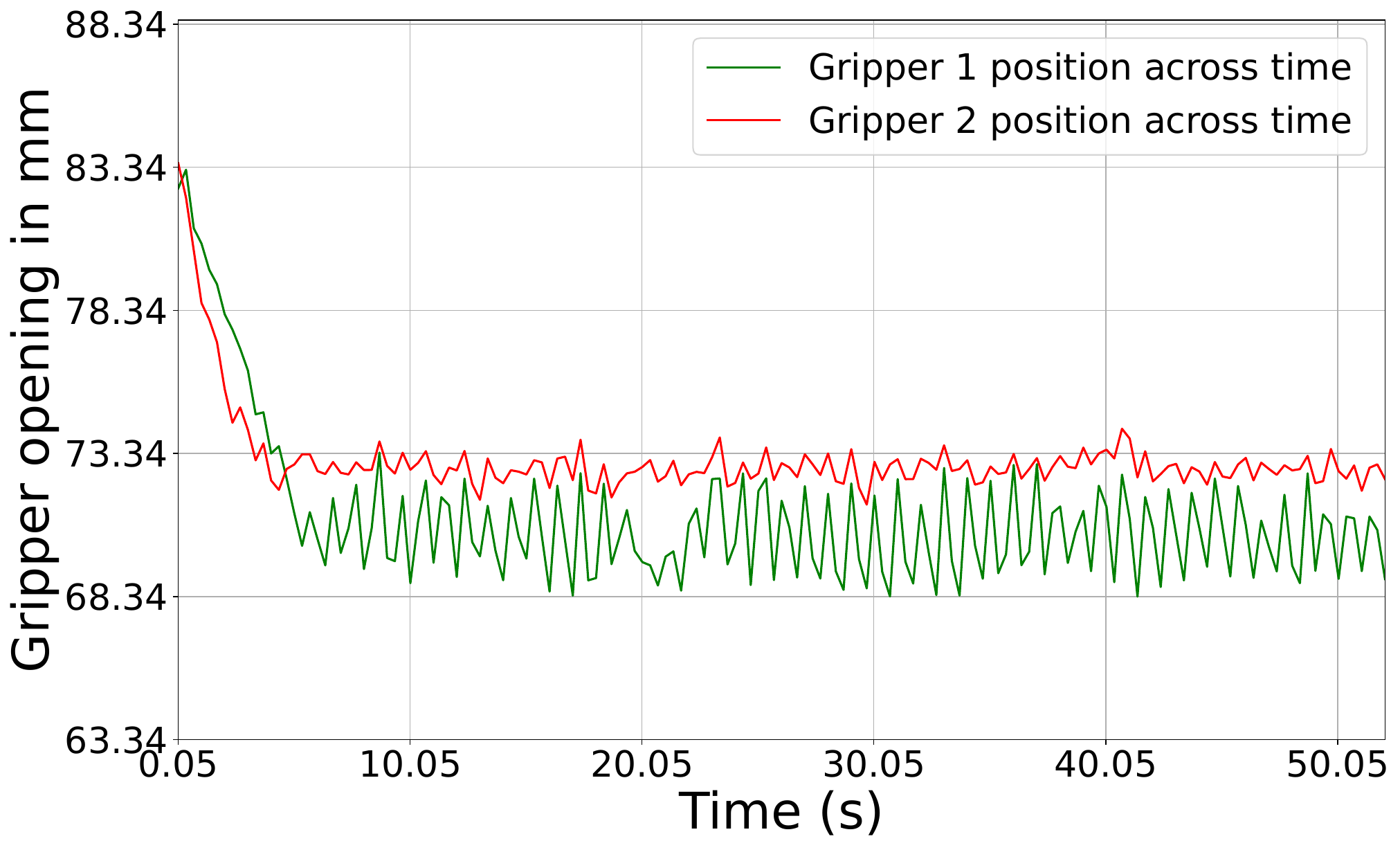} \\
\hline
\multicolumn{3}{c}{PVC Pipe} \\
\InclGr{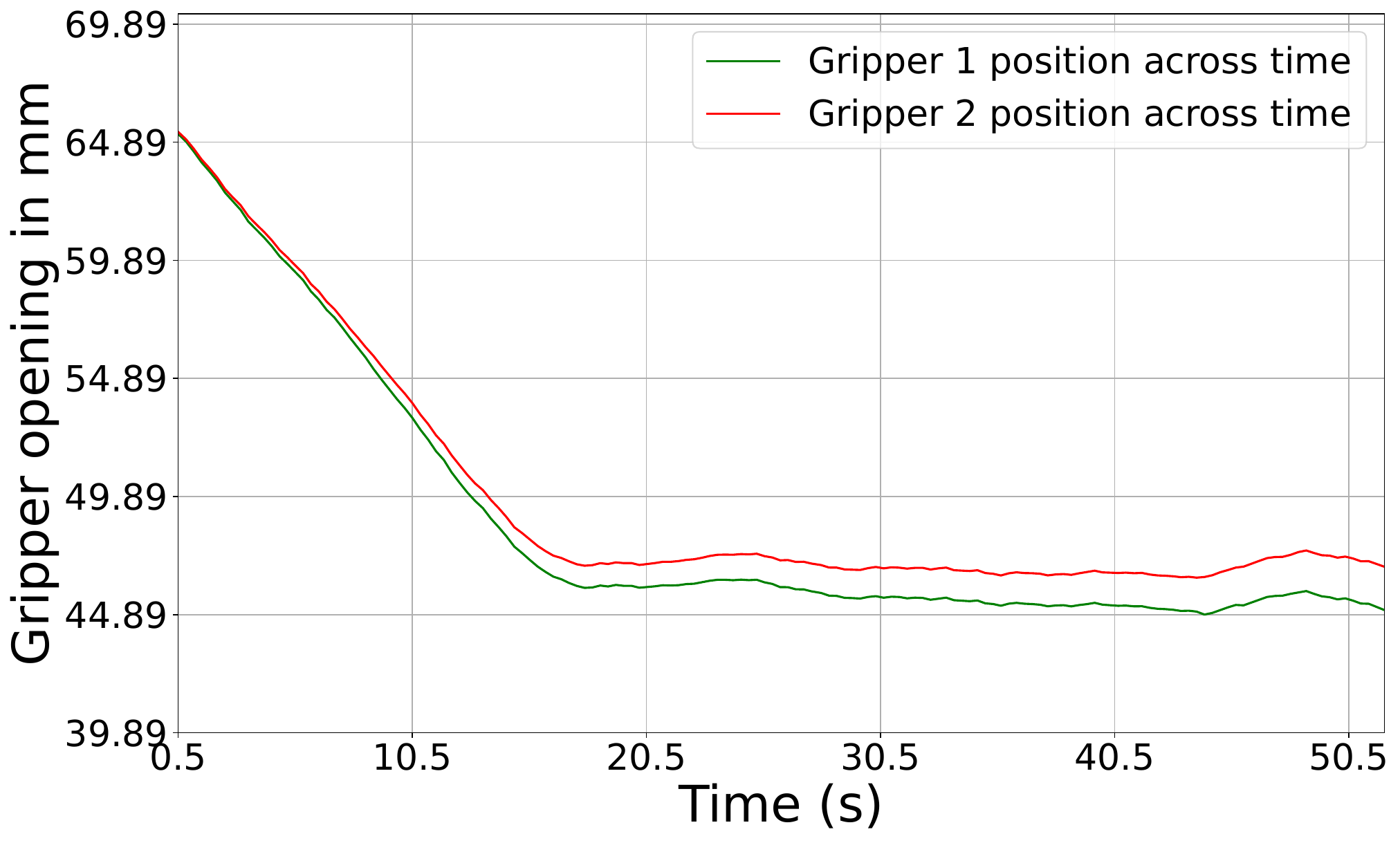} & 
\InclGr{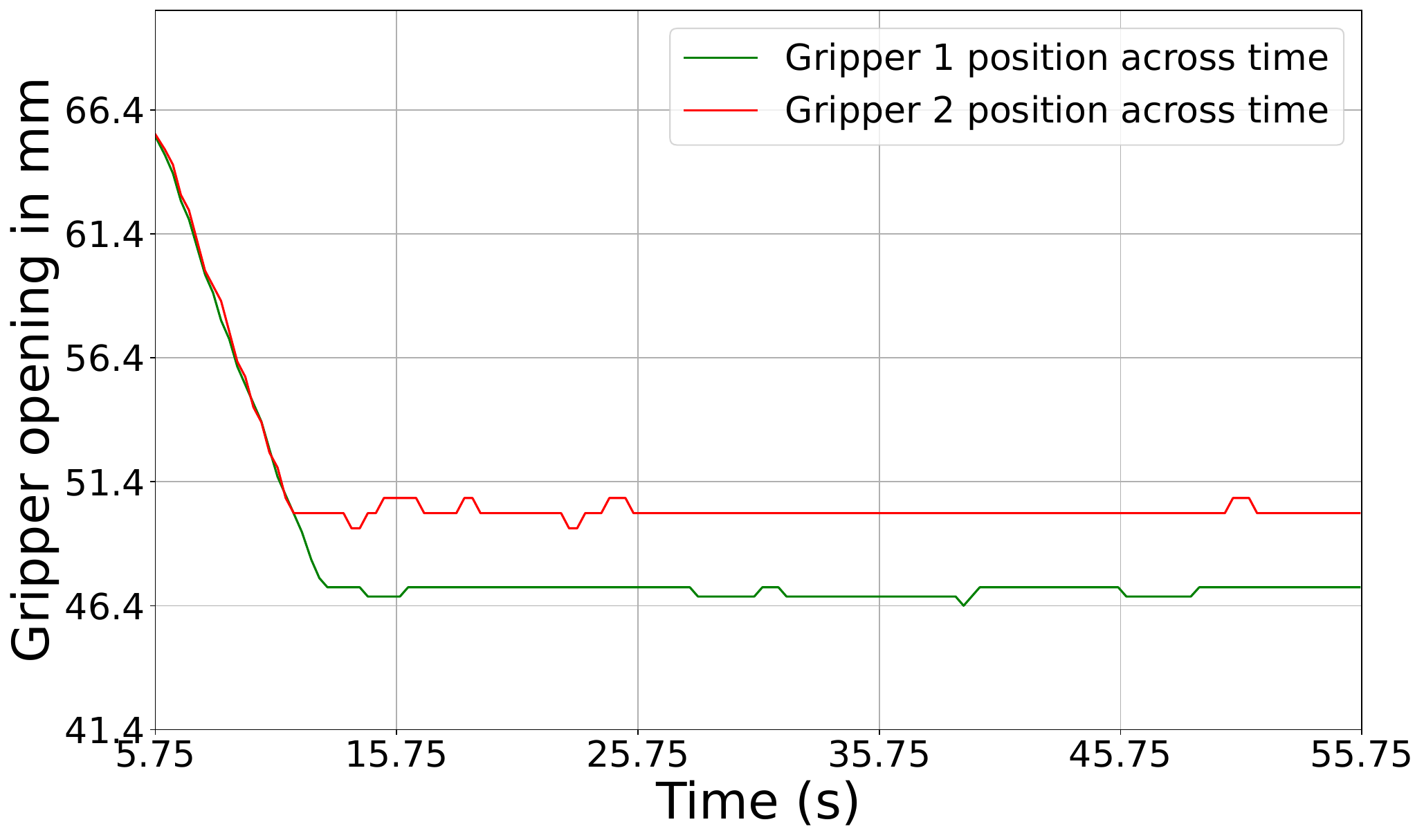} & 
\InclGr{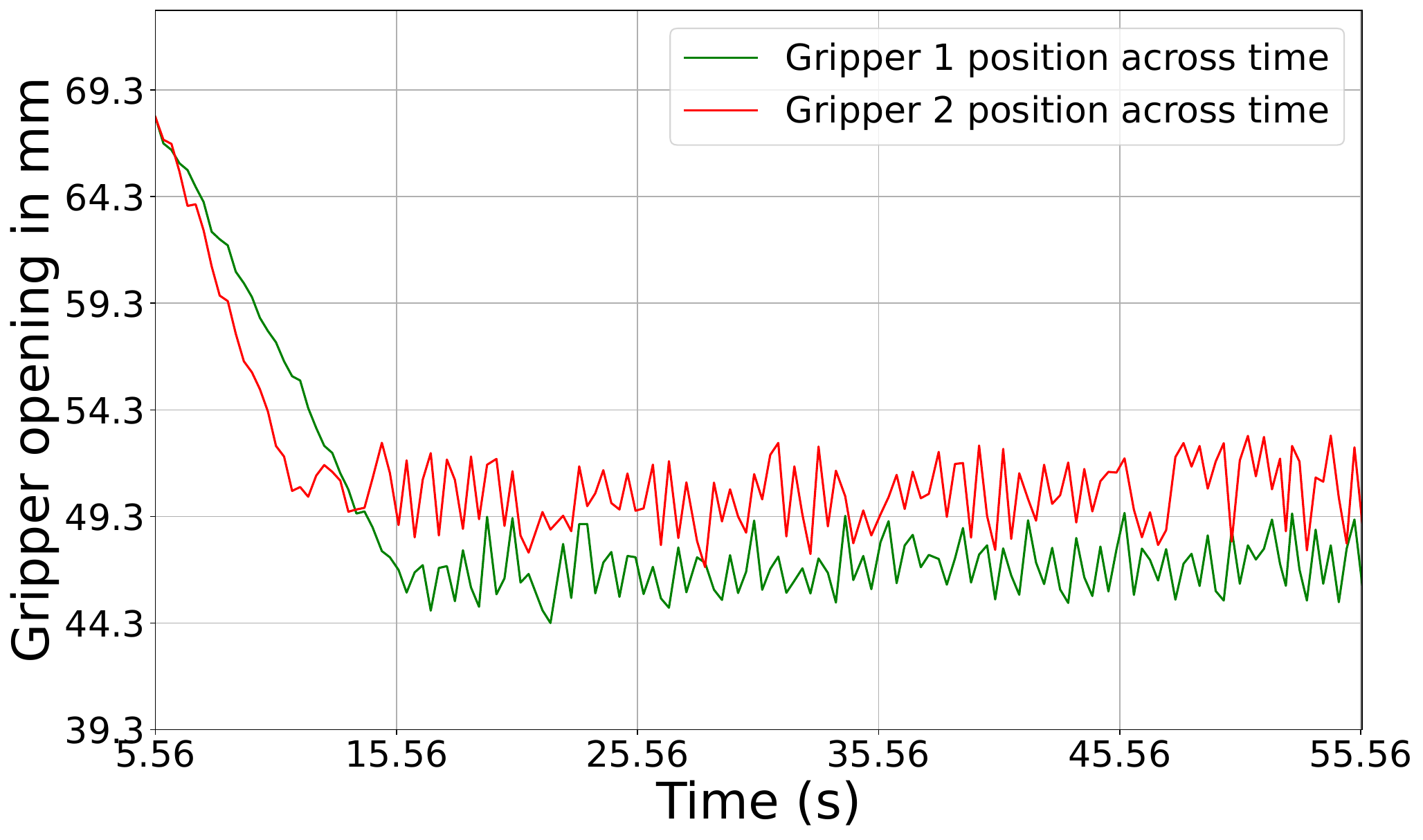} \\
\hline
\multicolumn{3}{c}{Coffee Beans Bag} \\
\InclGr{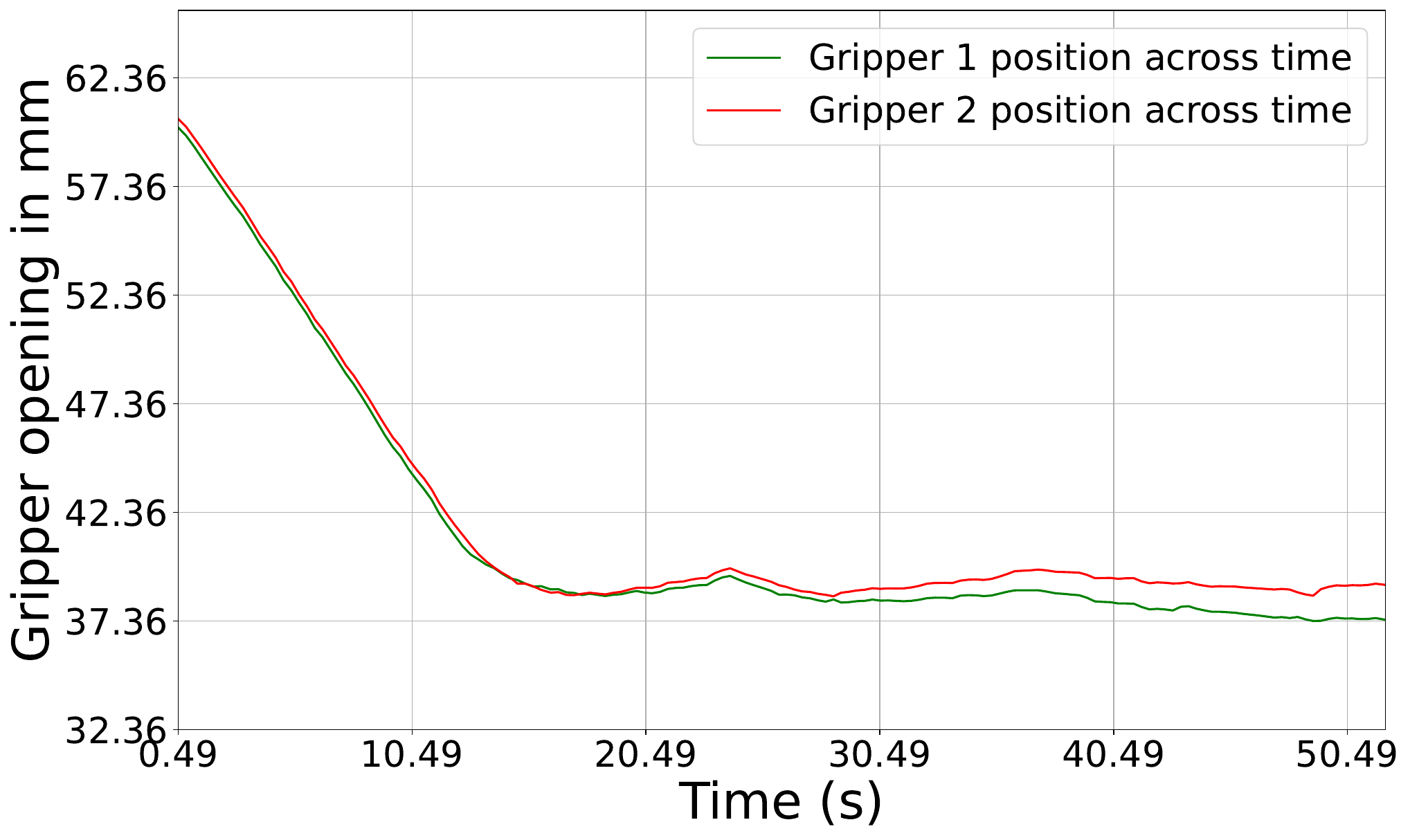} & 
\InclGr{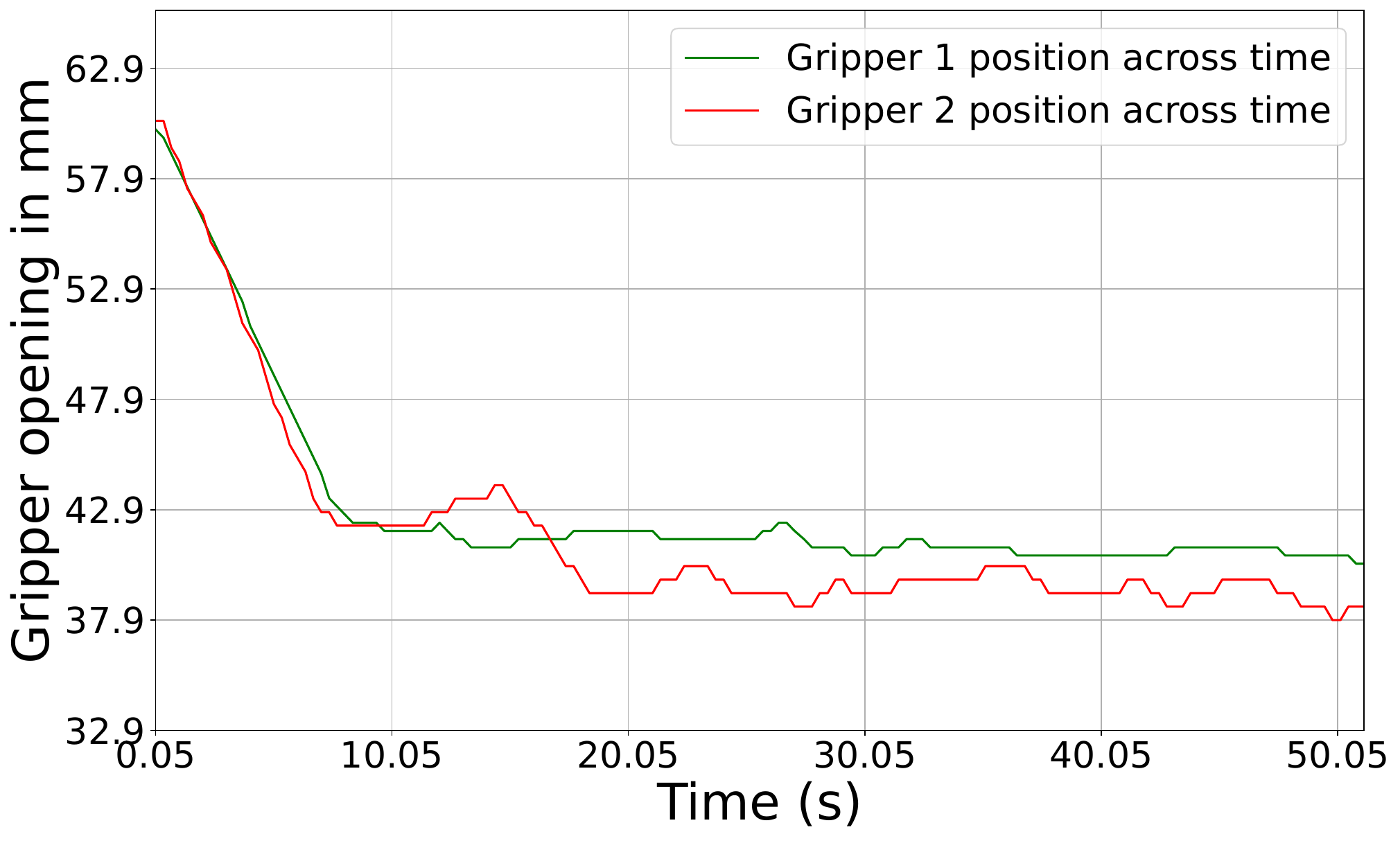} & 
\InclGr{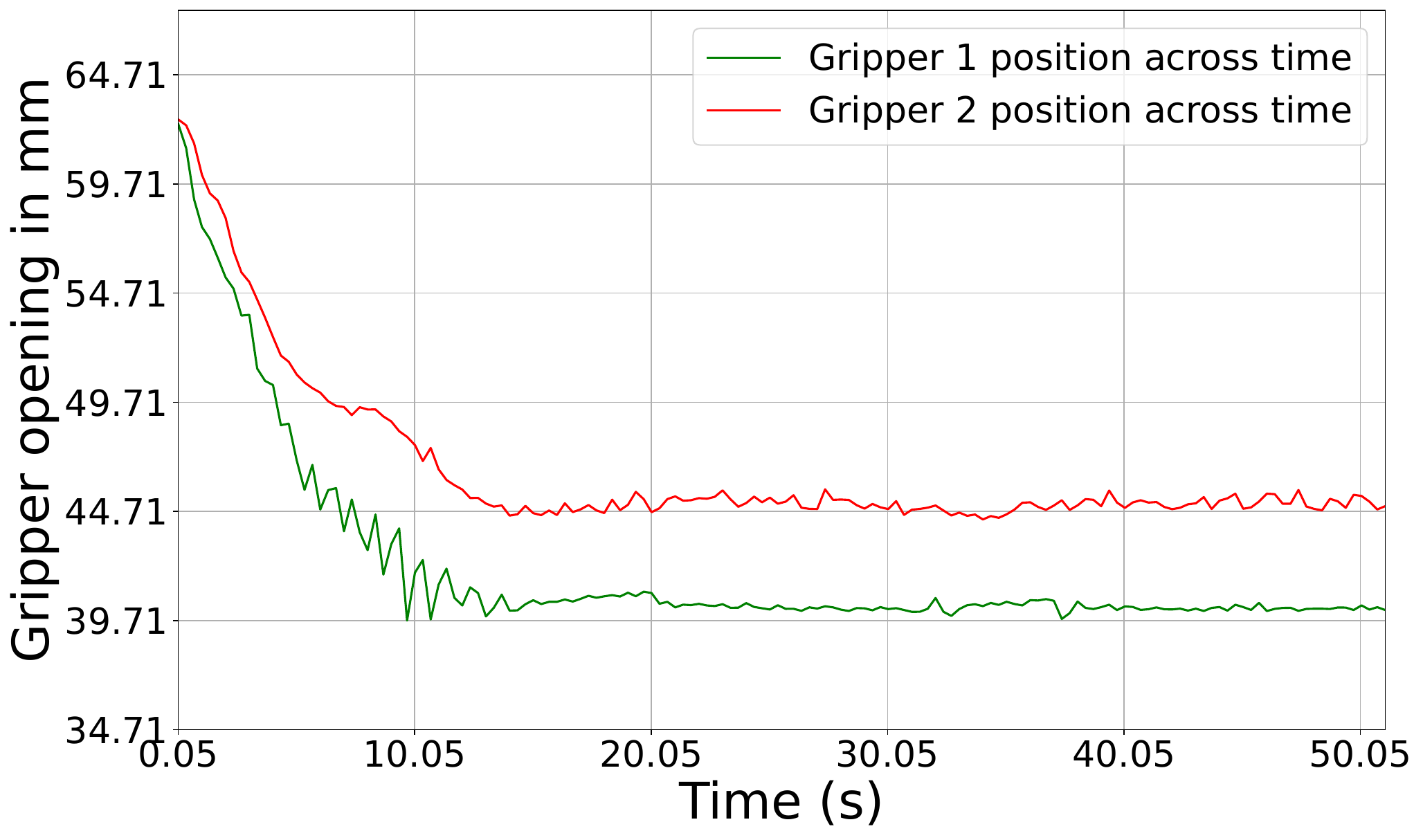} \\
\hline

\hline\hline
\end{tabular}
\endgroup
\caption{Model performance across objects.}
\label{fig:model_performance_across_objects}
\end{figure*}

\begin{figure}[h!]
    \centering
    \includegraphics[width=\linewidth]{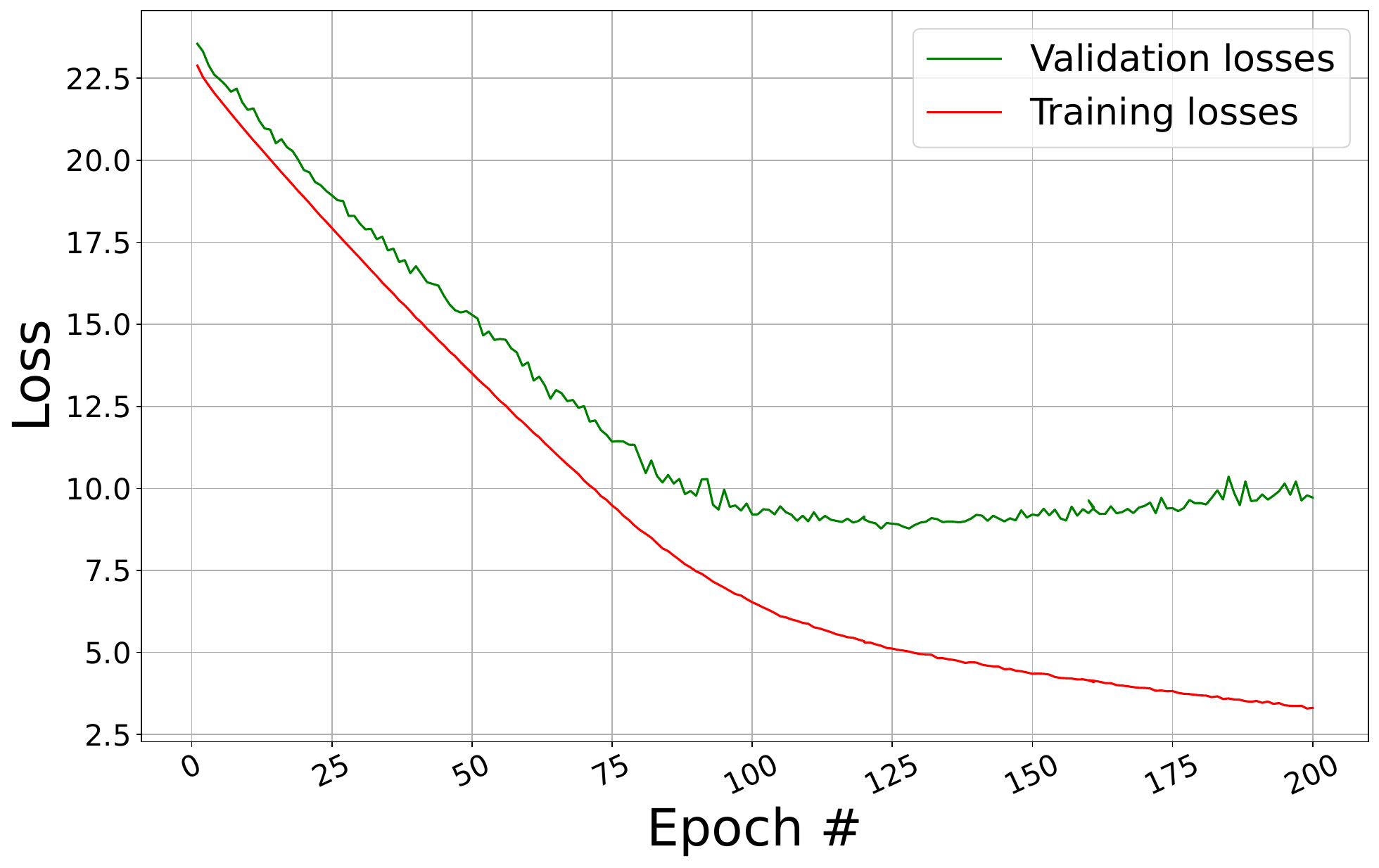}
    \caption{Losses vs. epoch.}
    \label{fig:pslip}
    \label{fig:training_loss}
\end{figure}

Three non-learned scalar parameters are also included: $Q_v$, $Q_a$, and $P_q$. These parameters influence the convergence behavior of the optimization layer but do not directly interact with the learned components of the system. Since their primary role is to regulate numerical stability and regularization, they can be manually tuned rather than learned. It is is critical that all three scalars remain strictly positive to ensure the stability and solvability of the optimization problem.

As discussed in the introduction, selecting appropriate values for the embedding size M and the prediction horizon N is critical to system performance. In our experiments, excessively large values of N led to high initial losses during training, often preventing the controller from converging. This behavior stems from the intrinsic nature of Model Predictive Control (MPC), which performs multi-step forecasting over the prediction horizon using learned and estimated dynamics. While the first few steps leverage real observations, subsequent predictions are based on recursive estimations. As the horizon length increases, compounding errors accumulate, leading to poor performance and inaccurate predictions at the later timesteps.

\begin{table}[h!]
\centering
\caption{MPC Layer Constants}
\label{tab:mpc_constants}
\begin{tabular}{ccccccc}
\toprule
$ \mathbf{P_q} $ & $\mathbf{Q_v}$ & $\mathbf{Q_a}$ & nHidden & nStep & $ \Delta t $ & $ \epsilon $ \\

\midrule
5 & 200 & 1 & 20 & 15 & 10 & 1e-4\\
\bottomrule
\end{tabular}
\end{table}

\begin{table}[h!]
\centering
\caption{CNN Encoder Constants}
\label{tab:cnn_encoder_constants}
\begin{tabular}{cccc}
\toprule
\text{hidden1} & \text{hidden2} & \text{embed\_dim} & \text{res\_size} \\
\midrule
128 & 128 & 20 & 224 \\
\bottomrule
\end{tabular}
\end{table}

\section{Results}

\renewcommand{\arraystretch}{2.0} 
\begin{table}[h!]
\centering
\caption{Success rate in real-world experiments}
\label{tab:trial_performance}
\begin{tabular}{cccc}
\hline
\textbf{Object Type} & 
\makecell{\textbf{Multi-Agent} \\ \textbf{MPC (Ours)}} & 
\makecell{\textbf{Single-Agent} \\ \textbf{MPC}} & 
\makecell{\textbf{Single-Agent} \\ \textbf{PD}} \\
\hline
Banana          & \textbf{10/10} & 9/10 & 7/10 \\ \hline
Pepperoni Stick & \textbf{10/10} & 7/10 & 6/10 \\ \hline
Pringles Can    & \textbf{10/10} & 8/10 & 5/10 \\ \hline
PVC Pipe        & \textbf{10/10} & \textbf{10/10} & 6/10 \\ \hline
Coffee Beans Bag& \textbf{9/10}  & 7/10  & 6/10 \\ \hline
\end{tabular}
\end{table}

We train the model for 200 epochs. To evaluate the advantages of the proposed layer, we focus on elongated objects with varying stiffnesses and compare the grasping success rate against baseline methods in Table \ref{tab:trial_performance}, with images of the respective objects in Figure \ref{fig:trial_objects}.

The experimental setup consists of five representative objects: a banana, pepperoni stick, Pringles can, PVC pipe, and a coffee bean bag. A trial is considered successful if the grippers converge to a stable grasp and maintain it without object slippage for a minimum of 15 seconds without incurring visible damage to the object being grasped. As evidenced in Figure \ref{fig:model_performance_across_objects}, even trials considered "successful" under the baseline methods exhibit pronounced fluctuations and inconsistent behavior across time. In contrast, our proposed multi-agent MPC layer achieves substantially smaller differences in opening between agents, resulting in markedly more stable grasps. Once the model converges to an appropriate opening, both agents maintain a steady grasp throughout the trial, due to the inherent collaborative nature of the layer. This visual comparison illustrates that, while baseline methods may barely achieve success, our approach provides superior stability. 

The LeTac-MPC baseline and our proposed model rely on the sensory data in the form of regular images, which are much more consistent across runs, making the replication of behavior more consistent for the controller. The baseline PD model utilizes a binary depth image as a reference to track. The depth sensors in most current tactile sensors are not of very high quality, resulting in an inability to maintain consistent values in the same conditions across multiple runs, making recalibration often necessary between runs. 

\section{Advantages Over Single-Agent Model}
Our proposed \textit{fully cooperative} MPC layer is one where both agents share a common goal (successful grasping and transportation of objects) and reward structure. This follows one of the most typical multi-agent model structures, where the success of one agent directly contributes to the success of the other(s) \cite{jin2025comprehensivesurveymultiagentcooperative}.

Our system addresses the limitations of single-agent grasping implementations by leveraging coordinated grasping from multiple manipulators. The presence of multiple contact points allows for significantly improved force distribution and object stabilization, which in turn increases the success rate of complex grasping tasks. The superiority of multi-agent grasping against single-agent implementations has been noted by implementations such as \cite{bernardtiong2024cooperativegraspingtransportationusing}. However, to the best of the author's knowledge, this is the first implementation of a centralized, multi-agent Model Predictive Control layer leveraging tactile sensing. Furthermore, our proposed MPC layer enables robust manipulation of objects with varying dimensions. Notably, the system is capable of maintaining stable grasps even when the agents exhibit asymmetric gripper openings, thereby reducing the risk of object slippage or failure during joint manipulation.

\renewcommand{\arraystretch}{2.0} 
\begin{table}[h!]
\centering
\caption{Runtime across batch sizes.
}
\label{tab:runtime_increase}
\begin{tabular}{cccc}
\hline
\textbf{\makecell{Batch \\ Size}} & \textbf{\makecell{Multi-Agent \\ RT (s)}} & \textbf{\makecell{Single-Agent \\ RT (s)}} & \textbf{\makecell{RT Increase \\ (\%)}} \\
\hline
1 & 0.025232 & 0.010666 & 136.56 \\ \hline
2 & 0.012691 & 0.012198 & 4.05 \\ \hline
4 & 0.019566 & 0.016118 & 21.39 \\ \hline
8 & 0.036248 & 0.025747 & 40.78 \\ \hline
16 & 0.068352 & 0.051429 & 32.90 \\ \hline
32 & 0.169282 & 0.112945 & 49.88 \\ \hline
64 & 0.330727 & 0.235298 & 40.56 \\ \hline
128 & 0.690623 & 0.481402 & 43.46 \\ \hline
\end{tabular}
\end{table}

Due to the homogeneous nature of our multi-agent framework, parameter sharing is employed to substantially enhance data utilization during training. This approach significantly reduces the volume of data required, as the parameters are learned simultaneously across both agents during their concurrent operation within the system.
As shown in Table \ref{tab:runtime_increase}, although our multi-agent implementation must process essentially twice the amount of input data and perform matrix operations with dimensions effectively doubled compared to the single-agent case, the observed runtime increase is only approximately 45\%. This represents a favorable computational trade-off, significantly outperforming the expectation of a 100\% slowdown under linear scaling assumptions. This suggests that the proposed multi-agent extension is not only effective but also computationally efficient, given the added benefits in task performance and reliability. It is worth noting that for batch sizes of one, the increased runtime is primarily attributed to computational overhead introduced by PyTorch, rather than inefficiencies in the core computation itself.

\section{Conclusion, Limitations, and Future Work}
In summary, our multi-agent tactile grasping approach leverages tactile RGB images rather than tactile depth images, which provides more consistent inputs and eliminates the need for frequent controller re-tuning. We demonstrate that the proposed controller generalizes effectively to previously unseen objects with varying sizes, stiffness, and surface textures. Comparative evaluations against baselines, including model-based MPC and PD controllers, further highlight the superior performance of our method. Overall, this project represents a notable advancement in multi-agent robotics, introducing the first framework to successfully integrate tactile sensing into robotic manipulation at this scale.

Despite these strengths, several limitations remain. First, tactile images vary noticeably across different GelSight sensors even under identical circumstances, and as a result, training on one sensor and deploying on another leads to a significant degradation in performance. Second, the controller is computationally more demanding than conventional model-based approaches, which may restrict its deployment in resource-constrained systems. Third, the framework struggles with very soft or deformable objects, which tend to collapse before meaningful tactile feedback is obtained. Finally, sharp or abrasive objects pose risks of permanently damaging the sensor.

Future work will address these limitations along several directions. Promising avenues include transfer and continual learning to improve cross-sensor robustness. Incorporating decentralized learning and imitation learning could further improve scalability and adaptability, while human-in-the-loop or human-robot collaboration frameworks may enhance the system's practical usability. Additionally, methods for mitigating sensor-to-sensor discrepancies remain an important step toward making this approach more widely deployable.

\bibliographystyle{IEEEtran}
\bibliography{sources}
\end{document}